\documentclass[preprint]{elsarticle}

\usepackage[utf8]{inputenc}  
\usepackage[T1]{fontenc}     
\usepackage{lmodern}         
\usepackage{microtype}       
\usepackage[scaled]{helvet} 
\usepackage[english]{babel}  
\usepackage{graphicx}        
\usepackage[bookmarks=false]{hyperref}            
\usepackage{algpseudocode}
\usepackage{algorithm}
\usepackage{comment}         
\biboptions{sort&compress}   
\usepackage{amsfonts}
\usepackage{amsthm}
\journal{xxxxx}
\usepackage{booktabs}
\usepackage{multirow}
\usepackage{multicol}
\newtheorem*{remark}{Remark}
\usepackage{amsmath,amsfonts,amssymb}

\usepackage{subfigure}

\usepackage{xcolor}
\usepackage{todonotes}
\usepackage[a4paper, total={7in, 9in}]{geometry}

\graphicspath{{fig/}}

\begin{document}

\begin{frontmatter}

\title{BP-DeepONet: A new method for cuffless blood pressure estimation using the physcis-informed DeepONet}


\author[1]{Lingfeng Li
}
\ead{lfli@hkcoche.org}

\author[2]{Xue-Cheng Tai\corref{cor1}}

\ead{xtai@norceresearch.no}

\author[1,3]{Raymond Hon-Fu Chan
}
\ead{raymond.chan@cityu.edu.hk}


\affiliation[1]{organization={Hong Kong Centre for Cerebro-cardiovascular Health Engineering},
city={Hong Kong},
country={China}}
\affiliation[2]{organization={Norwegian Research Center},
city={Bergen},
country={Norway}}
\affiliation[3]{organization={Department of mathematics, City university of Hong Kong},
city={Hong Kong},
country={China}}

\cortext[cor1]{Corresponding author}

\begin{abstract}
Cardiovascular diseases (CVDs) are the leading cause of death worldwide, with blood pressure serving as a crucial indicator. Arterial blood pressure (ABP) waveforms provide continuous pressure measurements throughout the cardiac cycle and offer valuable diagnostic insights. Consequently, there is a significant demand for non-invasive and cuff-less methods to measure ABP waveforms continuously. Accurate prediction of ABP waveforms can also improve the estimation of mean blood pressure, an essential cardiovascular health characteristic.
This study proposes a novel framework based on the physics-informed DeepONet approach to predict ABP waveforms. Unlike previous methods, our approach requires the predicted ABP waveforms to satisfy the Navier-Stokes equation with a time-periodic condition and a Windkessel boundary condition. Notably, our framework is the first to predict ABP waveforms continuously, both with location and time, within the part of the artery that is being simulated. Furthermore, our method only requires ground truth data at the outlet boundary and can handle periodic conditions with varying periods. Incorporating the Windkessel boundary condition in our solution allows for generating natural physical reflection waves, which closely resemble measurements observed in real-world cases. Moreover, accurately estimating the hyper-parameters in the Navier-Stokes equation for our simulations poses a significant challenge. To overcome this obstacle, we introduce the concept of meta-learning, enabling the neural networks to learn these parameters during the training process.
To demonstrate the effectiveness of our approach, we conduct numerical experiments that showcase its superiority over traditional methods, which typically only predict systolic blood pressure (SBP) and diastolic blood pressure (DBP). Our proposed method not only accurately predicts continuous ABP waveforms but also outperforms these traditional approaches.

\end{abstract}
\end{frontmatter}

\section{Introduction} 
According to the World Health Organization (WHO), cardiovascular diseases (CVDs) are the leading cause of death globally \cite{WHO}. In 2019, there were about 32\% of all global deaths caused by CVDs. The majority of these deaths were due to heart attack and stroke. There is sufficient evidence to support that high blood pressure (or hypertension) is the predominant cause of CVDs \cite{fuchs2020high}. In practice, the morphology of the arterial blood pressure (ABP) waveforms can reflect the cardiovascular status of humans \cite{matthieu2008cardiac,thiele2011arterial,li2017forward,hullender2021simulations}. Therefore, it is of great importance to monitor ABP waveforms continuously to detect and prevent diseases.

In the clinical setting, the gold standard for ABP waveform monitoring is the invasive method, which involves implanting an invasive pressure sensor into the artery. The invasive method is very risky, and the used devices are expensive. It is commonly used in intensive care units (ICUs). On the other hand, the non-invasive method is a more desirable choice for the daily monitoring of ABP waveforms.

\begin{figure}[h]
    \centering
    \subfigure[]{\includegraphics[width=0.4\textwidth]{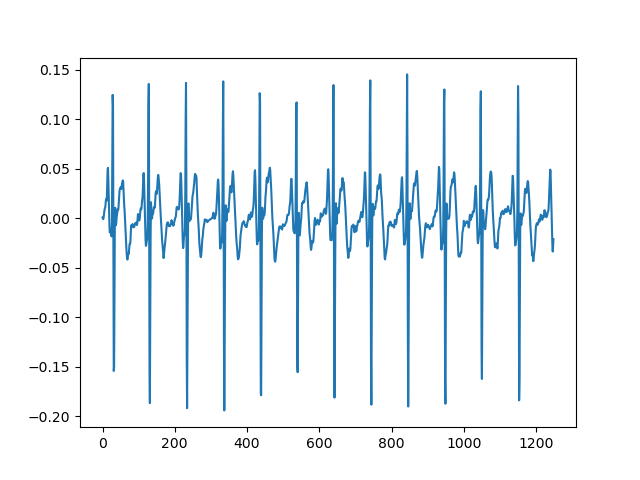}} 
    \subfigure[]{\includegraphics[width=0.4\textwidth]{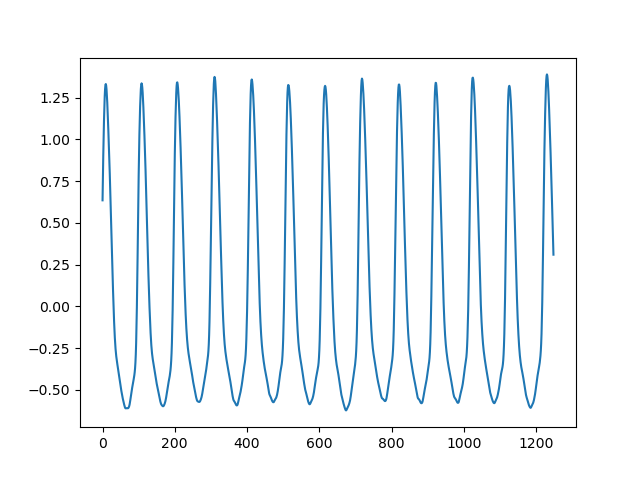}} 
    \caption{Filtered ECG and PPG signals in 10 seconds.}
    \label{fig:ecg_ppg}
\end{figure}

The deep learning-based method for non-invasive ABP waveform estimation has attracted huge research interest in recent years. These methods usually estimate the ABP from other physiological signals that are much easier to measure. Electrocardiogram (ECG) and photoplethysmogram (PPG) are the most commonly used signals. The ECG signal records the heart's electrical activity through cardiac cycles. It is measured by some electrodes, which can detect the small electrical changes resulting from the cardiac activity.
The patterns of ECG signals can indicate many cardiac abnormalities. The PPG signal is an optical signal that can detect blood volume changes in the microvascular bed of tissue. It is often obtained by illuminating the skin with a pulse oximeter and measuring changes in light absorption. Examples of filtered ECG and PPG signals are shown in Figure \ref{fig:ecg_ppg}.

Deep learning models were introduced to estimate systolic and diastolic blood pressure in \cite{su2018long}. Then, more complicated neural networks are used to estimate the entire ABP waveform \cite{eom2020end,ibtehaz2022ppg2abp,ma2022kd}. In \cite{eom2020end}, the authors proposed a neural network consisting of a convolutional neural network (CNN), a bi-directional gated recurrent unit (Bi-GRU), and attention layers to estimate ABP from ECG and PPG. In \cite{ibtehaz2022ppg2abp}, a network that predicts ABP only using PPG signals was proposed. In \cite{ma2022kd}, the authors introduced a modified transformer-based model to predict the ABP waveform from ECG, PPG, and some morphological extracted features.

\begin{figure}
    \centering
    \includegraphics[width=0.7\textwidth]{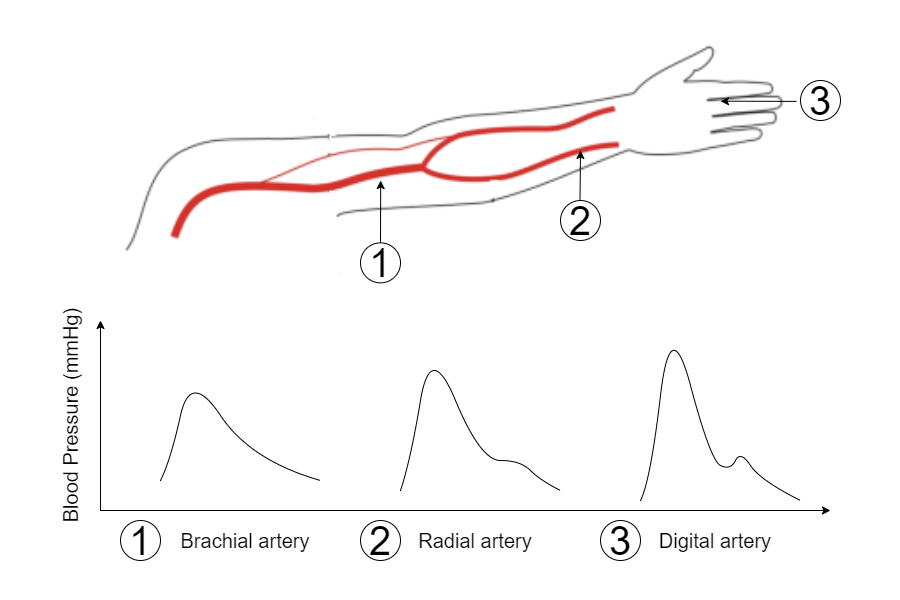}
    \caption{Illustration of the pulse pressure amplification in arterial systems.}
    \label{fig:pulseamplification}
\end{figure}

The models above are all trained in a data-driven way, i.e., trained by minimizing the difference between the ground truth label and the network prediction. Since the labeled data is usually measured by invasive methods in the radial artery \cite{lakhal2017invasive}, the trained neural network will only predict the radial ABP waveform. However, ABP waveforms differ at different locations in the arterial systems \cite{2008ContinuousNA,armstrong2019brachial} (See Figure \ref{fig:pulseamplification}), and radial waveforms may not be sufficient to diagnose CVD and reflect the status of the cardiovascular system. For example, some studies have shown that the brachial waveform is more valuable than the radial waveform when assessing cardiac contractility \cite{kyriazis2001dp,sharman2007radial}. Another important observation is the pulse pressure amplification effect. Pulse pressure is defined as the difference between the maximum and minimum of the pressure waveforms, and this difference will increase through the propagation of the pulse wave in the arterial system. All current deep learning-based algorithms can only predict radial ABP waveforms since they are trained with labeled data measured at the radial artery. 

In this work, we aim to develop new methods to predict ABP waveforms at different locations and at all times in the artery conditioned to the physical law that governs the blood flow, i.e. the Navier-Stokes equation. We use the physics-informed DeepONet (PI-DeepONet) \cite{wang2021learning}. DeepONet was initially proposed to learn the solution operator that maps variable inputs to the corresponding latent solutions of certain partial differential equation (PDE) systems \cite{lu2021learning}, and its training process is entirely data-driven. If the underlying PDE system is explicitly given, then the PDE can be incorporated into the training loss like the physics-informed neural networks (PINN) \cite{raissi2019physics}. The DeepONet trained in this way is called the physics-informed DeepONet. In the arterial system, the hemodynamics, i.e., blood flow dynamics, can be modeled and simulated by the Navier-Stokes equation \cite{formaggia2003one,quarteroni2004mathematical,xiao2014systematic}. Therefore, we can incorporate the Navier-Stokes equation into the training loss and learn the solution operator that maps physiological signals to solutions of the corresponding PDEs. In our setting, the spatial domain of the Navier-Stokes equation is a segment of the artery, and the labeled data can serve as a Dirichlet boundary condition. After providing other necessary initial/boundary conditions, we can identify the PDE solution in the whole domain. Consequently, we can obtain the ABP waveforms at different locations and at all times.

We summarize our main contributions as follows:
\begin{itemize}
    \item This work is the first attempt to apply physics-informed operator learning to ABP waveform estimation. The proposed methods are the first ones to predict ABP waveforms at different locations in the artery conditioned by the physical law that governs the blood flow, i.e., the Navier-Stokes equation. The proposed methods enforce neural network solutions to satisfy a 1-D Navier-Stokes equation with a Windkessel boundary condition and a time-periodic condition. The Windkessel boundary condition can simulate physical reflection waves. 
    \item The proposed methods can predict the blood pressure waveforms with reasonably good accuracy and preserve some physical properties like pulse pressure amplification.
    \item To the best of our knowledge, the proposed methods are the first ones that incorporate variable time-periodic conditions to DeepONet. We also propose an effective implementation method for the network training.
    \item We also incorporate meta-learning into the training to estimate model hyper-parameters automatically from the input physiological signals. 
    \item During training, we only require the blood pressure measurement at the outlet side of the domain, and no extra measurement is needed.  
\end{itemize}

\section{One-dimensional model for blood flow simulation}

\begin{figure}
    \centering
    \includegraphics[width=\textwidth]{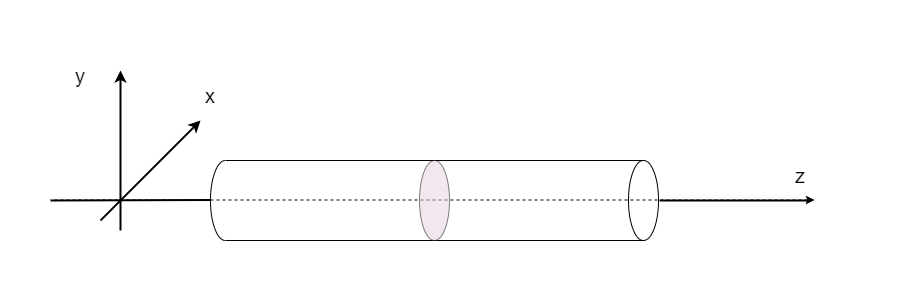}
    \caption{Simplification of the three-dimensional model to one-dimensional model. The vessel is assumed to be a straight cylinder, and each axial cross-section is circular.}
    \label{fig:vessel}
\end{figure}

The hemodynamics in a vessel segment can be effectively modeled and simulated by the one-dimensional Navier-Stokes equation \cite{quarteroni2004mathematical,formaggia2003one}. 
The one-dimensional model is a simplification of the three-dimensional Navier-Stokes equation by making several assumptions on the computational domain:
\begin{itemize}
    \item The vessel is a straight cylinder with the axis oriented along the coordinate $z$ direction (See Figure \ref{fig:vessel}).
    \item The vessel wall displaces along the radial direction only, which implies that each axial cross-section is always circular.
    \item The blood pressure and blood flow rate are constant within each axial cross-section.
    \item All body forces, like the gravity, are neglected.
\end{itemize}

Thus, the hemodynamics can be described by three quantities, namely the blood flow rate $Q(z,t)$, the axial cross-sectional area $A(z,t)$, and the blood pressure $P(z,t)$, where $z\in[0,L]$ is the spatial coordinates and $t\in[0,T]$ is the temporal coordinates. $L$ denotes the length of the vessel segment, and $T$ represents the time duration.
Then, the one-dimensional model can be written as:
\begin{align}
    & \frac{\partial A}{\partial t}+\frac{\partial Q}{\partial z}=0, \label{eq:NS_1} \\
    & \frac{\partial Q}{\partial t}+\frac{\partial}{\partial z}(\frac{Q^2}{A})+\frac{A}{\rho}\frac{\partial P}{\partial z}+K_r\frac{Q}{A}=0, \label{eq:NS_2}
\end{align}
where $\rho$ is the blood density, and $K_r$ is a resistance parameter related to blood viscosity. To close this system, the blood pressure $P$ and the cross-sectional area are assumed to satisfy:
\begin{equation}
    P=P_{ext}+\beta\frac{\sqrt{A}-\sqrt{A_0}}{A_0}, \label{eq:A_P}
\end{equation}
where $P_{ext}$ is the external pressure, $A_0$ is the cross-sectional area at the reference state, and $\beta$ is a constant depending on the physical and mechanical properties of vessels. More details of the model derivation can be found in \cite{quarteroni2004mathematical,formaggia2003one}. A simple characteristic analysis shows that the forward characteristic variable $W_1$ and backward characteristic variable $W_2$ in the system (\ref{eq:NS_1})-(\ref{eq:NS_2}) are:
$$ W_1(A,Q)=\frac{Q}{A}+4\sqrt{\frac{\beta}{2\rho A_0}}(A^{1/4}-A_0^{1/4}),$$
and
$$ W_2(A,Q)=\frac{Q}{A}-4\sqrt{\frac{\beta}{2\rho A_0}}(A^{1/4}-A_0^{1/4}). $$
The forward wave and backward wave are traveling at the speed 
$$\lambda_1(A,Q)=\frac{Q}{A}+\sqrt{\frac{\beta}{2\rho A_0}}A^{1/4}$$ 
and 
$$\lambda_2(A,Q)=\frac{Q}{A}-\sqrt{\frac{\beta}{2\rho A_0}}A^{1/4}$$ 
respectively. In practice, we have
$$\sqrt{\frac{\beta}{2\rho A_0}}A^{1/4}\gg Q/A, $$ 
so $W_1$ and $W_2$ always travel in opposite directions. The speed of the forward wave $\lambda_1\approx \sqrt{\frac{\beta}{2\rho A_0}}A^{1/4}$ is often called pulse wave velocity. 

Using the relation (\ref{eq:A_P}), we can rewrite the equations (\ref{eq:NS_1})-(\ref{eq:NS_2}) as
\begin{equation}
    \frac{\partial U}{\partial t}+H(U)\frac{\partial U}{\partial z}=B(U),\quad (z,t)\in[0,L]\times [0,T] \label{eq:NS_system}
\end{equation}
where
\begin{equation*}
    U=\begin{pmatrix}
        P \\ Q
    \end{pmatrix},
\end{equation*}
\begin{equation*}
    H(U)=\begin{pmatrix}
        0 & \frac{\beta}{2a(P)A_0} \\ \frac{a(P)^2}{\rho}-\frac{2A_0Q^2}{\beta a^3(P)} & \frac{2Q}{a(P)^2}
    \end{pmatrix},
\end{equation*}
\begin{equation*}
    B(U)=\begin{pmatrix}
        0 \\ -K_r\frac{Q}{a(P)^2}
    \end{pmatrix},
\end{equation*}
and
\begin{equation*}
    a(P)=\frac{A_0}{\beta}(P-P_{ext})+\sqrt{A_0}.
\end{equation*}

Even though the one-dimensional model can not capture the blood flow details, it can effectively describe the wave propagation within the vessel. This model has been validated by both \textit{in vitro} data \cite{alastruey2011pulse,saito2011one} and \textit{in vivo} data \cite{olufsen2000numerical,steele2003vivo}. There are many effective numerical solvers, like the MacCormack scheme\cite{elad1991numerical,fullana2009branched}, Taylor-Galerkin scheme \cite{wan2002one,sherwin2003computational}, MUSCL \cite{cavallini2008finite,delestre2013well}, and the local discontinuous Galerkin scheme \cite{matthys2007pulse,marchandise2009numerical,mynard20081d}. A comparison of different numerical methods has also been done in \cite{wang20141d}.

\section{Physics-informed machine learning for PDEs}
Machine learning algorithms for PDEs have been extensively studied in recent years. In general, there are two main directions in this research field: learning an instance of PDEs and learning the solution operator of PDEs. The most famous work along the first direction is the physics-informed neural networks (PINNs) proposed in \cite{raissi2019physics}. This work aims to solve an instance of PDEs using a neural network. Compared to traditional numerical solvers, the PINN is mesh-free, easy to implement, and can solve high-dimensional problems. However, it can not achieve very high accuracy for non-linear PDEs. The second direction aims to learn the solution operator that maps from some input spaces to the solution space using some neural networks. This type of network is also called the neural operator. The most popular neural operators include the DeepONet \cite{lu2021learning} and the Fourier Neural operator (FNO) \cite{li2020fourier}. The learned network can quickly predict the PDE solution corresponding to the given input. We will provide a general introduction to two types of methods in this section.

\subsection{Physics-informed neural networks (PINNs)}
Let's consider a general form of a $d$-dimensional boundary value problem:
\begin{align*}
    \mathcal{L}(u)(x)&=f(x),\quad x\in\Omega\\
    \mathcal{B}(u)(x)&=g(x),\quad x\in\partial\Omega
\end{align*}
where $\Omega\in\mathbb{R}^d$ is the problem domain, $\mathcal{L}(u)$ denotes a general differential operator, and $\mathcal{B}(u)$ denotes corresponding boundary conditions. The PINN method first defines a neural network to approximate the solution to this specific instance of PDE: $\mathcal{N}(x;\theta):\Omega\rightarrow\mathbb{R}$, where $\theta$ denotes the set of all trainable parameters in the network. Then, this network is trained by solving the optimization problem:
\begin{equation*}
    \min_\theta  \sum_{j=1}^{N_0} |\mathcal{L} (\mathcal{N}(x_j;\theta))-f(x_j)|^2 + \omega\sum_{j=1}^{N_1}|\mathcal{B}(\hat{x}_{j})-g(\hat{x}_{j})|^2
\end{equation*}
where $\omega$ is the weight of the boundary residual, $\{x_{j}\}_{j=1}^{N_0}$ is collocation points randomly sampled from $\Omega$, and $\{\hat{x}_{j}\}_{j=1}^{N_1}$ is collocation points randomly generated from $\partial\Omega$.
The partial derivatives of $\mathcal{N}$ can be easily calculated through the back-propagation mechanism. An efficient implementation is provided in the DeepXDE package \cite{lu2021deepxde} for both PyTorch and TensorFlow platforms. Some similar algorithms in the literature use similar ideas, such as the deep Ritz method (DRM) \cite{yu2018deep} and the mixed residual method (MIM) \cite{lyu2022mim, yang2021local}. The DRM uses the variational formulation of PDEs to train networks, and the MIM transforms high-order PDEs into first-order systems to train networks. Various numerical analyses have also been conducted for these physics-informed methods \cite{lu2021priori, muller2021error,mishra2022estimates,li2022generalization, li2022priori}.

\begin{figure}
    \centering
    \subfigure[Fully connected network]{\includegraphics[width=0.9\textwidth]{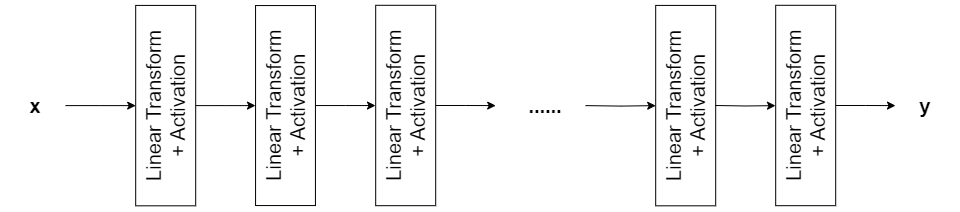}} 
    \subfigure[Residual network]{\includegraphics[width=0.9\textwidth]{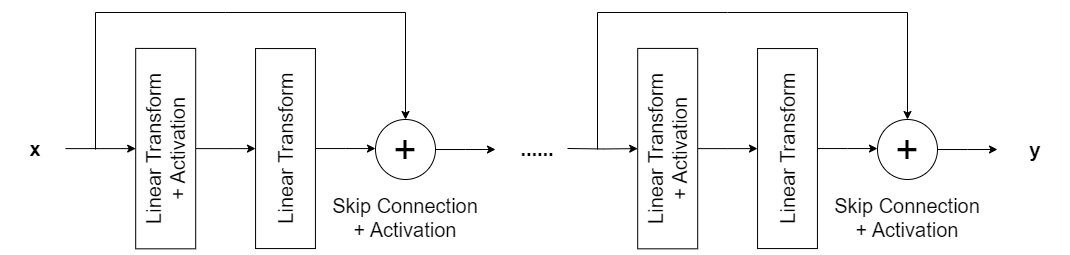}} 
    \caption{Structures of the fully connected network and the residual network.}
    \label{fig:structure1}
\end{figure}

One popular network structure for the PINN method is the fully connected network (FCN) (Figure \ref{fig:structure1}(a)). An FCN is typically defined as compositions of many fully connected layers: 
\begin{align*}
    &\mathcal{N}_{FCN}(x;\theta):=(\mathcal{F}_L\circ\mathcal{F}_{L-1}\circ\dots\circ\mathcal{F}_1)(x),\\
    &\mathcal{F}_l(z):=\sigma_l(W_lz+b_l),\quad \forall l=1,\dots,L-1, \\
    & W_l\in\mathbb{R}^{d_{l,in}\times d_{l,out}},\quad b_l\in\mathbb{R}^{d_{l,out}},\\
    &d_{1,in}=d,\quad d_{l,in}=d_{l-1,out},\quad l=2,\dots,L-1, \\
    &\mathcal{F}_L(z):=\sigma_L(W_Lz+b_L),\quad W_L\in\mathbb{R}^{d_{L-1,out},1},\quad b_L\in\mathbb{R},
\end{align*}
where $\sigma_l$ is the activation function for each layer, $d_{l,in}$ denotes the input dimension of the $l$th layer, and $d_{l,out}$ denotes the output dimension of the $l$th layer. Choices for the activation function usually include Sigmoid, Tanh, ReLU, etc. For the last layer $F_L$, the activation function can be the identify function, i.e., $\sigma_L(z)=z$. When increasing the depth $L$, the training of the FCN may become slower because of the vanishing gradient effect. To overcome this limitation, we can use the residual network incorporating the skip connection structure \cite{he2016deep}. A residual network (Figure \ref{fig:structure1}(b)) is defined as:
\begin{align*}
    &\mathcal{N}_{Res}(x;\theta):=(\hat{\mathcal{F}}_L\circ\hat{\mathcal{F}}_{L-1}\circ\dots\circ\hat{\mathcal{F}}_1)(x),\\
    &\hat{\mathcal{F}}_l(z):=\sigma_l(W_{l,2}\sigma_l(W_{l,1}z+b_{l,1})+b_{l,2}+z),\quad \forall l=1,\dots,L-1,\\
    &W_{l,1}\in\mathbb{R}^{d_{l,in}\times d_{l,out}},\quad W_{l,2}\in\mathbb{R}^{d_{l,out}\times d_{l,out}},\quad b_{l,1},b_{l,2}\in\mathbb{R}^{d_{l,out}},\\
    &d_{1,in}=d,\quad d_{l,in}=d_{l-1,out},\quad l=2,\dots,L-1,\\
    &\mathcal{F}_L(z):=\sigma_L(W_Lz+b_L),\quad W_L\in\mathbb{R}^{d_{L-1,out},1},\quad b_L\in\mathbb{R}.
\end{align*}

\subsection{Neural operators for PDEs}


Let's consider a class of parametric PDEs:
\begin{align*}
    \mathcal{L}(u)(x;\alpha)&=f(x;\alpha),\quad x\in\Omega\\
    \mathcal{B}(u;\alpha)(x)&=g(x;\alpha), \quad x\in\partial\Omega,
\end{align*}
where $\alpha$ is the parameters that define the problem; for example, $\alpha$ can be the diffusion coefficient in a convection-diffusion equation. The neural operator aims to approximate the solution operator $\mathcal{M}:\mathcal{A}\rightarrow\mathcal{U}$, where $\mathcal{A}$ is the function space for $\alpha$, and $\mathcal{U}$ is the space for the solution $u$. In practice, the input space can also be other function spaces, e.g., the space of the source term $f(x)$. Two types of neural operators have been extensively applied to various tasks in the literature: the DeepONet \cite{lu2021learning} and the FNO \cite{li2020fourier}. 

\begin{figure}
    \centering
    \subfigure[DeepONet]{\includegraphics[width=0.45\textwidth]{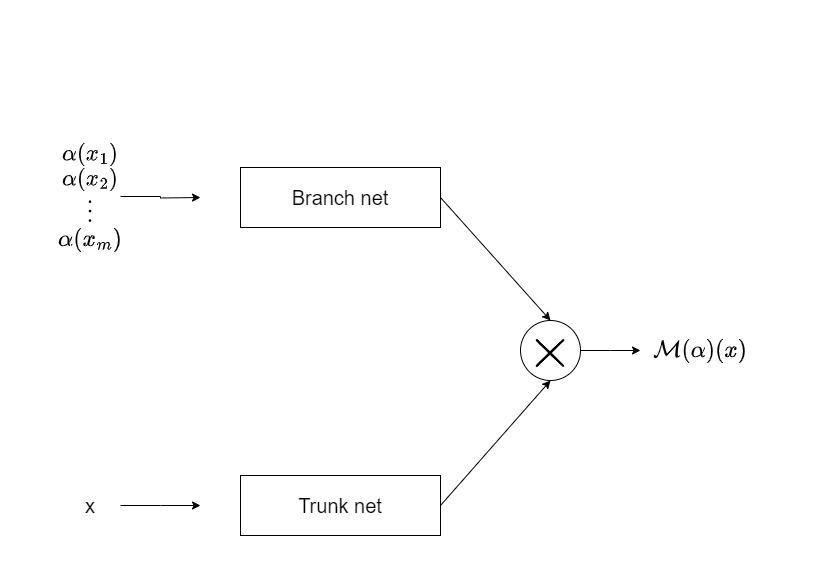}} 
    \subfigure[FNO]{\includegraphics[width=0.45\textwidth]{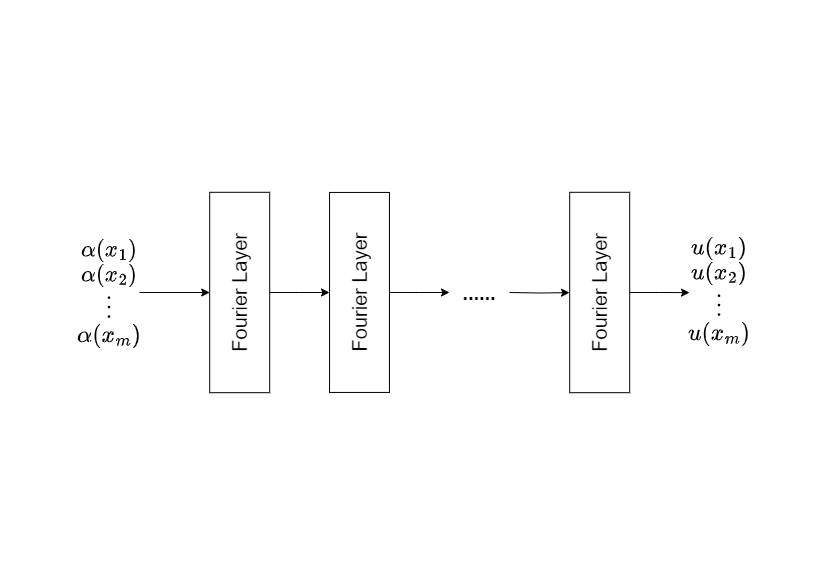}} 
    \caption{Structures of the DeepONet and FNO.}
    \label{fig:structure2}
\end{figure}

A plain DeepONet (Figure \ref{fig:structure2}(a)) consists of two sub-networks: a branch net and a trunk net. It is defined as
\begin{equation*}
    \mathcal{N}_{DON}(\hat{\alpha},x):=Branch(\hat{\alpha})^\top Trunk(x)
\end{equation*}
where $\hat{\alpha}$ is the discretization of $\alpha$ on a given mesh, $x$ is any point in $\Omega$. The outputs of the branch net and the trunk net are $p$-dimensional vectors, where $p$ is a user-specified hyper-parameter. The DeepONet $\mathcal{N}(\hat{\alpha},x)$ gives a prediction to $\mathcal{M}(\alpha)(x)$. To predict the solution on multiple points, we need to evaluate the trunk net multiple times.

A plain FNO (Figure \ref{fig:structure2}(b)) is the composition of many Fourier layers:
\begin{align*}
    &\mathcal{N}_{FNO}(\hat{\alpha}):=(\Tilde{\mathcal{F}}_L\circ\Tilde{\mathcal{F}}_{L-1}\dots\circ \Tilde{\mathcal{F}}_1)(\hat{\alpha})\\
    &\tilde{\mathcal{F}}_l(z):=\sigma_l(W_lz+FFT^{-1}(R_l(FFT(z))))
\end{align*}
where $\sigma_l$ is the activation function, $FFT$ is the fast Fourier transformation, $FFT^{-1}$ is the inverse fast Fourier transformation, $W_l$ is a linear transformation, and $R_l(z)$ is an operator that includes a truncation of first $m$ frequency modes and mode-wise complex matrix multiplications. The output of the FNO is of the same size as $\hat{\alpha}$.  

The training of neural operators is fully data-driven. Suppose we have a set of $\{\hat{\alpha}_i\}_{i=1}^N$ from $\mathcal{A}$ and corresponding PDE solutions $\{\hat{u}_i\}_{i=1}^N$ defined on a discretization $\{x_j\}_{j=1}^M$. Then we train the neural operator by minimizing the $L_2$ difference between the network outputs and the reference solution:
\begin{equation*}
    \min_\theta \frac{1}{MN}\sum_{i=1}^N\sum_{j=1}^M |\mathcal{N}_{DON}(\hat{\alpha}_i,x_j)-\hat{u}_i(x_j)|^2
\end{equation*}
for the DeepONet, or 
\begin{equation*}
    \min_\theta \frac{1}{N}\sum_{i=1}^N|\mathcal{N}_{FNO}(\hat{\alpha}_i)-\hat{u}_i|_2^2
\end{equation*}
for the FNO. A comprehensive study of two different neural operators is given in  \cite{lu2022comprehensive}, which includes numerical experiments on various linear and non-linear PDEs. In general, the performances of two types of neural operators are comparable in various problems. The FNO is more efficient in training and testing, while the DeepONet has a more flexible structure that can be easily applied to different tasks. 

Besides the data-driven training method, we can apply the physics-informed loss like PINN to train neural operators. Such methods are called physics-informed neural operators \cite{wang2021learning, li2021physics, goswami2022physics}. In many engineering problems, obtaining the entire solution to PDEs may be very difficult or expensive. Consequently, it is more desirable to incorporate the physics laws into the training process to regularize the neural operators. We can construct a similar residual loss as the PINN method for each pair of training samples and minimize the summation over all samples. For the DeepONet, we can use back-propagation to evaluate the derivatives. For the FNO, we can use numerical differentiation to evaluate the derivatives. 

\section{The physics informed DeepONet for ABP waveform estimation}

\subsection{Problem setup}
This work aims to learn the operator that maps the physiological signals to the solution of the Navier-Stokes equation:
\begin{equation*}
    \mathcal{O}:\mathcal{S}([0,T])\rightarrow \mathcal{U}([0,L]\times [0,T]) \label{eq:operator}
\end{equation*} 
where $\mathcal{S}([0,T])$ is the set of physiological signals of length $T$ and $\mathcal{U}([0,L]\times [0,T])$ is the space of solutions to Navier-Stokes equations defined on $[0,L]\times [0,T]$.

Suppose we have a training dataset consisting of $N$ pairs of samples: 
$$\{(s_i,p_i)\}_{i=1}^N,$$ 
where $s_i$ is a set of physiological signals, and $p_i$ is the corresponding ABP waveform measured in the radial artery that lies in the forearm. All waveform data have the same length. The physiological signal $s_i$ serves as the input to the neural network and the blood pressure waveform $p_i$ is used to define the boundary condition. 


For each sample $i=1,\dots,N$, its hemodynamics is defined by $U_i=(P_i,Q_i)$. We assume each $U_i$ satisfies the Navier-stokes system described in (\ref{eq:NS_system}):
\begin{equation}
    \frac{\partial U_i}{\partial t}+H(U_i)\frac{\partial U_i}{\partial z}=B(U_i),\quad (z,t)\in[0,L]\times [0,T] \label{eq:NS_system_sampleindependent}.
\end{equation}

\begin{figure}
    \centering
    \subfigure[Prescribed inlet boundary condition]{\includegraphics[width=0.3\textwidth]{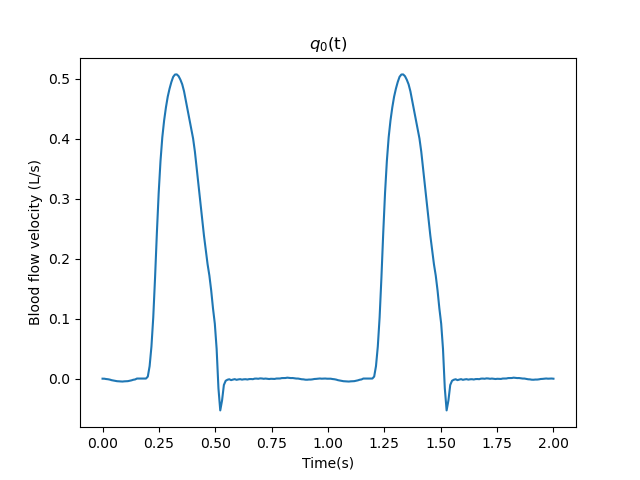}} 
    \subfigure[Simulated blood pressure using the non-reflecting boundary condition]{\includegraphics[width=0.3\textwidth]{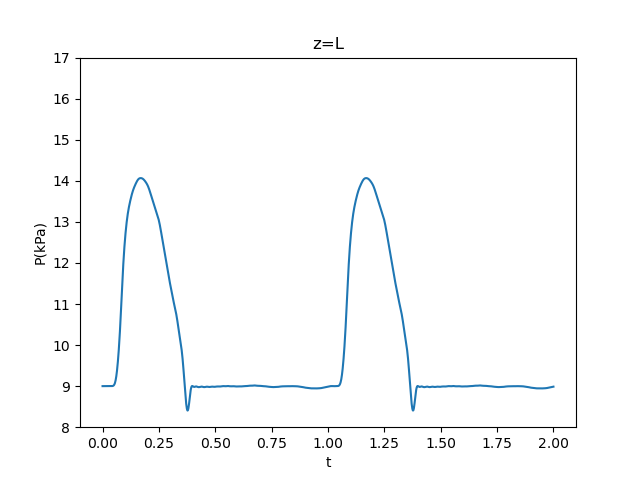}} 
    \subfigure[Simulated blood pressure using the Windkessel boundary condition]{\includegraphics[width=0.3\textwidth]{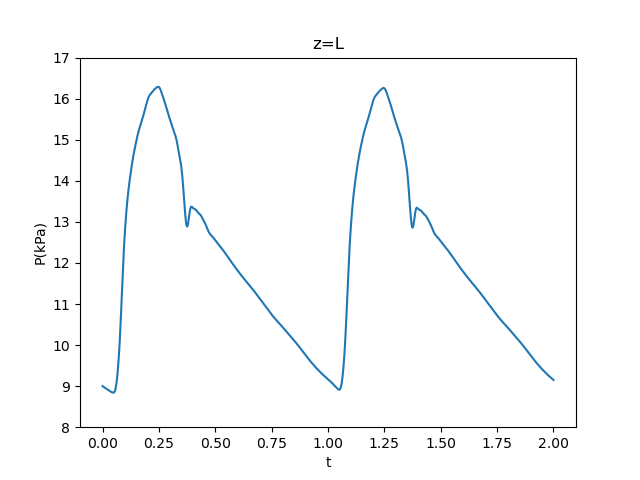}}
    \caption{Comparison of the simulated blood pressure using different outflow boundary conditions. (a) is the inlet boundary condition given at $z=0$; (b) is the simulated blood pressure at $z=L$ using the non-reflecting outflow boundary condition; and (c) is the simulated blood pressure at $z=L$ using the 3-element Windkessel outflow boundary condition. (b) and (c) are computed using a standard MacCormack scheme.}
    \label{fig:compare_bc}
\end{figure}

We also need to provide suitable initial and boundary conditions to identify unique PDE solutions. When solving this system using some traditional numerical methods, we usually need to impose an inlet boundary condition and an outlet boundary condition. The inlet boundary condition is usually set as a Dirichlet boundary condition which prescribes the value of the blood pressure $P$ or blood flow rate $Q$ at the inlet side. The outlet boundary condition can determine the reflection waves that run backward in the artery. A simple choice for the outlet boundary condition is the non-reflecting boundary condition \cite{quarteroni2004mathematical}: $W_2=0$ where $W_2$ is the backward characteristic variable. Though the non-reflecting boundary condition is very easy to implement, it can not provide a physically correct reflecting wave. A better choice is the three-element Windkessel boundary condition \cite{westerhof2009arterial}:
\begin{equation}
    W(P_i,Q_i):=Q_i(L,t)(1+\frac{R_{1}}{R_{2}})+CR_{1}\frac{\partial Q_i}{\partial t}(L,t)-\frac{P_i(L,t)}{R_{2}}-C\frac{\partial P_i}{\partial t}(L,t)=0,\quad t\in[0,T] \label{eq:Wind_bc}.
\end{equation} 
where $R_{1}$, $R_{2}$, and $C$ are some hyper-parameters related to the artery properties. 
\begin{remark}
    The Windkessel model describes the hemodynamics of the arterial system in terms of resistance and compliance, and it is widely used as an outflow boundary condition in the numerical simulation of human arterial systems \cite{alastruey2011pulse,alastruey2012arterial,xiao2014systematic}.  The Navier-Stokes equation with this Windkessel boundary condition can produce physical reflection waves in the solutions. A simple comparison of two outflow boundary conditions is shown in Figure \ref{fig:compare_bc} where we can observe that the blood pressure simulated using the non-reflecting boundary condition is not physically correct.
\end{remark}

In our case, since the pressure data $p_i$, $i=1,\dots,N$, is measured near the end of radial arteries, it is better to formulate it also as an outlet boundary condition:
\begin{equation}
    P_i(L,t)=p_i(t),\quad t\in[0,T] \label{eq:Dirchlet_bc}.
\end{equation}
To better simulate the reflection waves in the artery, we use the three-element Windkessel model (\ref{eq:Wind_bc}) as another boundary condition. 

For the initial condition, we impose a time-periodic condition on both $P_i$ and $Q_i$ instead of giving a prescribed initial value:
\begin{equation}
    U_i(z,t+\delta_i)=U_i(z,t),\quad (z,t)\in[0,L]\times[0,T] \label{eq:periodic}
\end{equation}
where $\delta_i$ is the period for the $i$-th training sample, and it can be estimated from the physiological signals $s_i$. Typically, the waveforms are almost periodic within a short time. 

\begin{remark}
    Conventional methods directly prescribe the initial value of solutions. One commonly used initial condition is $U(z,0)=(P_{ext},0)$. However, it is very difficult to measure the value of $P_i$ and $Q_i$ at the initial time for each sample and it is not realistic to assume an initial condition like $U(z,0)=(P_{ext},0)$. Therefore, using a time-periodic condition is more suitable in our situation.
\end{remark}

By solving the system (\ref{eq:NS_system}) with the boundary condition (\ref{eq:Dirchlet_bc}) (\ref{eq:Wind_bc}) and initial condition (\ref{eq:periodic}), we would be able to predict the hemodynamics in the radial artery. For conventional numerical methods, it would be difficult to solve under this setting, because there is no prescribed initial value given and there is no boundary condition at the inlet side; but it is easy to handle by the physics-informed method. 

In this problem, the period for each sample is different, which will cause some difficulty in computing the loss function. In this work, we will propose a new efficient implementation method to handle this issue.

\subsection{BP-DeepONet}\label{sec:BP-DeepONet}
\begin{figure}
    \centering
    \includegraphics[width=1\textwidth]{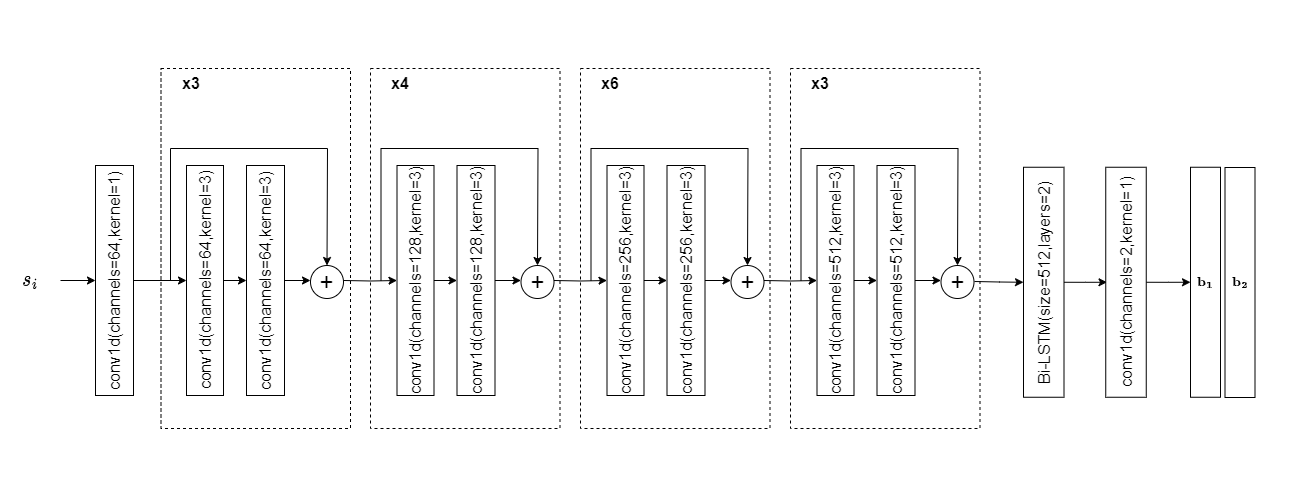}
    \caption{Structure of the branch net.}
    \label{fig:branch}
\end{figure}

\begin{figure}
    \centering
    \includegraphics[width=1\textwidth]{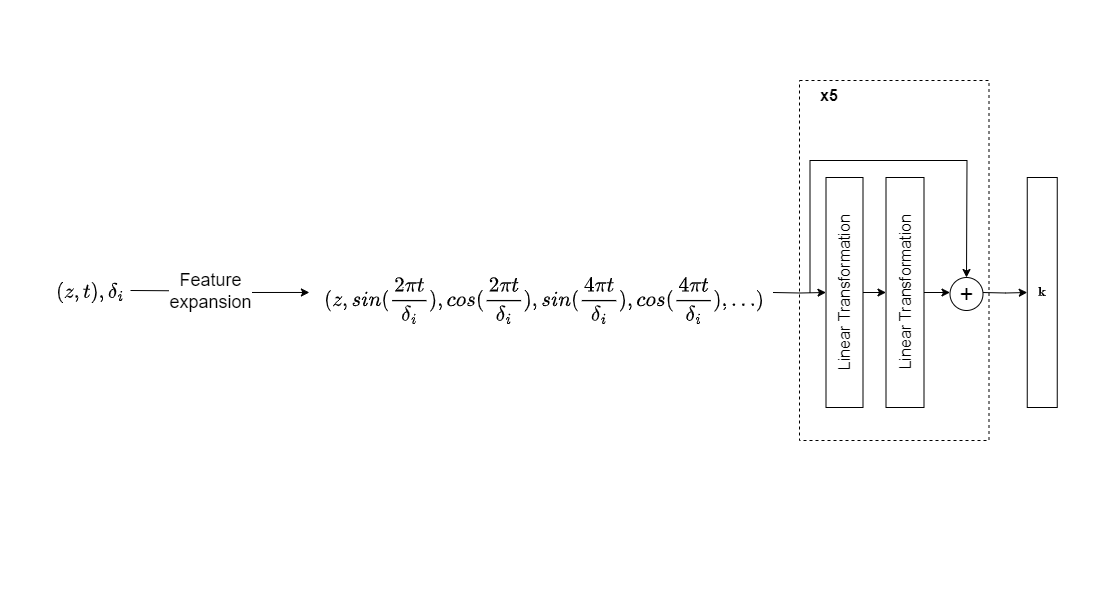}
    \caption{Structure of the trunk net.}
    \label{fig:trunk}
\end{figure}

We propose a new physics-informed DeepONet, BP-DeepONet, to learn the operator (\ref{eq:operator}). We adopt the same structure as the original DeepONet. For the branch net, we use a structure that combines a one-dimensional convolutional ResNet and a bi-directional LSTM, see Figure \ref{fig:branch}. The inputs to the branch net are some segments of physiological signals $s_i$ and the output would be two latent vectors $\mathbf{b_1}$ and $\mathbf{b_2}$. conv1d(channels=$c$,kernel=$s$) represents a standard one-dimensional convolutional layer whose number of output channels is $c$ and kernel size is $s$. Bi-LSTM (size=$h$,layers=$l$) represents a bi-directional LSTM block whose input size and hidden size are $h$, and the number of layers is $l$. A similar structure has already been used in \cite{paviglianiti2022comparison} to extract features from physiological signals to predict ABP waveforms. For the trunk net, we adopt a simple, fully connected ResNet as Figure \ref{fig:structure1}(b) with the feature expansion (Figure \ref{fig:trunk}). Feature expansion \cite{lu2021physics} is a technique to impose the periodic condition on PINN. To impose the time-periodic constraints (\ref{eq:periodic}), we can simply extend the inputs $(z,t)$ to
$$ (z,\sin(\frac{2\pi t}{\delta_i}),\cos(\frac{2\pi t}{\delta_i}),\sin(\frac{4\pi t}{\delta_i}),\cos(\frac{4\pi t}{\delta_i}),\dots). $$
Then, our network outputs will be strictly time-periodic with period $\delta_i$. The output of the trunk net is a vector $\mathbf{k}$ whose dimension is the same as $\mathbf{b_1}$ and $\mathbf{b_2}$. The final prediction of the BP-DeepONet will be calculated as follows:
$$ P_\theta(s_i,\delta_i,(z,t))=\mathbf{b}_1(s_i)^\top\mathbf{k}(z,t,\delta_i),\quad Q_\theta(s_i,\delta_i,(z,t))=\mathbf{b}_2(s_i)^\top\mathbf{k}(z,t,\delta_i) $$
where $\theta$ is the set of all trainable parameters in the BP-DeepONet.

\begin{figure}
    \centering
    \includegraphics[width=1\textwidth]{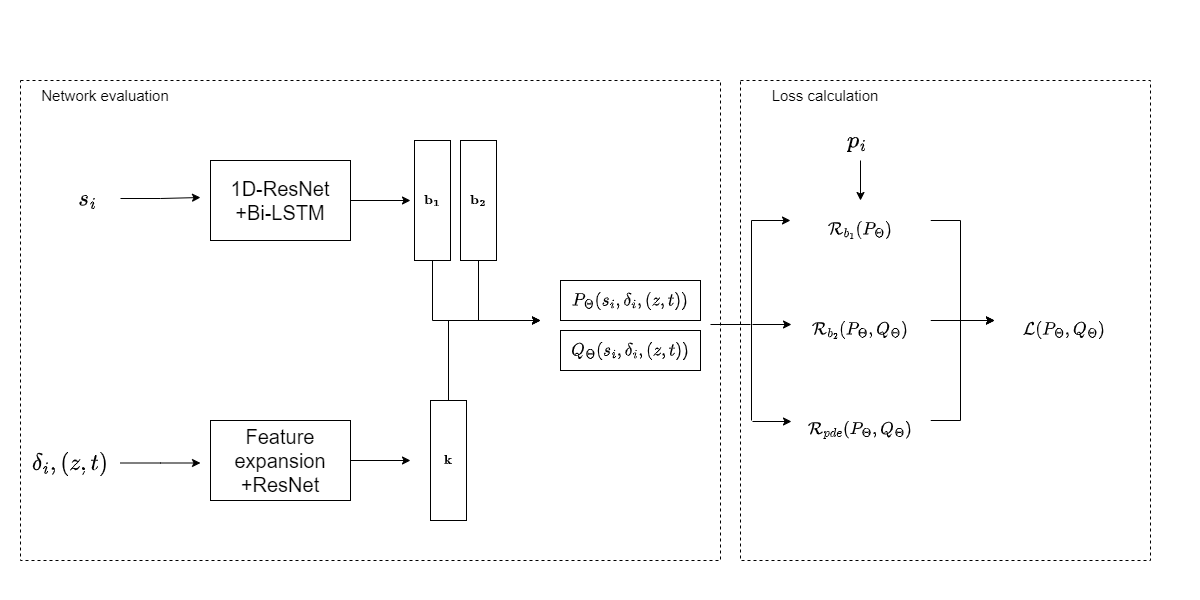}
    \caption{Structure of the BP-DeepONet}
    \label{fig:BP-DeepONet}
\end{figure}

The training loss function of our network is defined as
\begin{equation}
    \mathcal{L}(P_\theta,Q_\theta)=\mathcal{R}_{pde}(P_\theta,Q_\theta)+\omega_1\mathcal{R}_{b_1}(P_\theta)+\omega_2\mathcal{R}_{b_2}(P_\theta,Q_\theta) \label{eq:training_loss}
\end{equation}
where 
\begin{equation*}
    \mathcal{R}_{pde}(P_\theta,Q_\theta)=\frac{1}{N}\sum_{i=1}^N\left\Vert \left(\frac{\partial}{\partial t}\begin{pmatrix}P_\theta\\Q_\theta\end{pmatrix}+H(P_\theta,Q_\theta)\frac{\partial}{\partial z}\begin{pmatrix}P_\theta\\Q_\theta\end{pmatrix}-B(P_\theta,Q_\theta)\right)(s_i,\delta_i,\cdot)\right\Vert^2_{L^2([0,L]\times[0,T])},
\end{equation*}
\begin{equation*}
    \mathcal{R}_{b_1}(P_\theta)=\frac{1}{N}\sum_{i=1}^N\left\Vert P_\theta(s_i,\delta_i,(L,\cdot))-p_i\right\Vert_{L^2([0,T])}^2,
\end{equation*}
and
\begin{equation*}
    \mathcal{R}_{b_2}(P_\theta,Q_\theta)=\frac{1}{N}\sum_{i=1}^N\left\Vert W(P_\theta,Q_\theta)(s_i,\delta_i,(L,\cdot))\right\Vert_{L^2([0,T])}^2.
\end{equation*}
The complete structure of the BP-DeepONet is displayed in Figure \ref{fig:BP-DeepONet}.
To evaluate the different residual terms $\mathcal{R}_{pde}$, $\mathcal{R}_{b1}$, and $\mathcal{R}_{b2}$, we need to approximate the $L_2$ norm with some discrete quadrature. For the PDE residual term $\mathcal{R}_{pde}$, we can simply approximate the $i$-th $L_2$ residual by
\begin{align*}
    &\left \Vert \left(\frac{\partial}{\partial t}\begin{pmatrix}P_\theta\\Q_\theta\end{pmatrix}+H(P_\theta,Q_\theta)\frac{\partial}{\partial z}\begin{pmatrix}P_\theta\\Q_\theta\end{pmatrix}-B(P_\theta,Q_\theta)\right)(s_i,\delta_i,\cdot)\right\Vert^2_{L^2([0,L]\times[0,T])} \\
    \approx & \sum_{j=1}^M \left|\left(\frac{\partial}{\partial t}\begin{pmatrix}P_\theta\\Q_\theta\end{pmatrix}+H(P_\theta,Q_\theta)\frac{\partial}{\partial z}\begin{pmatrix}P_\theta\\Q_\theta\end{pmatrix}-B(P_\theta,Q_\theta)\right)(s_i,\delta_i,(z_j,t_j))\right|^2,
\end{align*}
where $\{(z_j,t_j)\}_{j=1}^M$ is a set of quadrature points randomly sampled from the domain $[0,L]\times[0,T] $.
The partial derivatives in the residual terms can be calculated as 
$$ \frac{\partial}{\partial x} P_\theta(s_i,\delta_i,(z_j,t_j))=\mathbf{b_1}(s_i)^\top\frac{\partial}{\partial x} \mathbf{k}(\delta_i,(z_j,t_j)),\quad x=z \text{ or } t, $$
$$ \frac{\partial}{\partial x} Q_\theta(s_i,\delta_i,(z_j,t_j))=\mathbf{b_2}(s_i)^\top\frac{\partial}{\partial x} \mathbf{k}(\delta_i,(z_j,t_j)),\quad x=z \text{ or } t. $$
However, evaluating this approximation is very time-consuming because the total number of gradient evaluations in one training epoch is $O(MN)$. It is necessary to design a more efficient method to approximate the residuals. This can be achieved by utilizing the periodicity of our BP-DeepONet. 

\subsection{Efficient implementation of training BP-DeepONet}
Given any periodic function $f(t)$ with period $\delta$, we have the norm $\Vert\cdot\Vert_{L^2([0,T])}$ is equivalent to $\Vert\cdot\Vert_{L^2([0,\delta])}$: 
$$ \left\lfloor \frac{T}{\delta}\right\rfloor\Vert f\Vert^2_{L^2([0,\delta])}\leq \Vert f\Vert^2_{L^2([0,T])}\leq \left\lceil \frac{T}{\delta}\right\rceil\Vert f\Vert^2_{L^2([0,\delta])}, $$
where $\lfloor\cdot\rfloor$ and $\lceil\cdot\rceil$ denote the floor function and ceiling function, respectively. 
Then, for the $i$-th $L_2$ residual term in $\mathcal{R}_{pde}$, we may replace it with
\begin{align*}
    &\left \Vert \left(\frac{\partial}{\partial t}\begin{pmatrix}P_\theta\\Q_\theta\end{pmatrix}+H(P_\theta,Q_\theta)\frac{\partial}{\partial z}\begin{pmatrix}P_\theta\\Q_\theta\end{pmatrix}-B(P_\theta,Q_\theta)\right)(s_i,\delta_i,\cdot)\right\Vert^2_{L^2([0,L]\times[0,\delta_i])} \\
    = & \int_0^{\delta_i}\int_0^L \left| \left(\frac{\partial}{\partial t}\begin{pmatrix}P_\theta\\Q_\theta\end{pmatrix}+H(P_\theta,Q_\theta)\frac{\partial}{\partial z}\begin{pmatrix}P_\theta\\Q_\theta\end{pmatrix}-B(P_\theta,Q_\theta)\right)(s_i,\delta_i,(z,t))\right|^2dzdt\\
    = & \int_0^{\delta_i}\int_0^L \left| \left(\frac{\partial}{\partial t}\begin{pmatrix}P_\theta\\Q_\theta\end{pmatrix}+H(P_\theta,Q_\theta)\frac{\partial}{\partial z}\begin{pmatrix}P_\theta\\Q_\theta\end{pmatrix}-B(P_\theta,Q_\theta)\right)(s_i,1,(z,t/\delta_i))\right|^2dzdt.
\end{align*}
After a change of variable $\hat{t}=t/\delta_i$, the above integration can be written as
\begin{align*}
    & \int_0^{1}\int_0^L \left| \left(\frac{\partial}{\partial \hat{t}}\begin{pmatrix}P_\theta\\Q_\theta\end{pmatrix}/\delta_i+H(P_\theta,Q_\theta)\frac{\partial}{\partial z}\begin{pmatrix}P_\theta\\Q_\theta\end{pmatrix}-B(P_\theta,Q_\theta)\right)(s_i,1,(z,\hat{t}))\right|^2dzd\hat{t}\notag\\
    \approx & \sum_{j=1}^M \left| \left(\frac{\partial}{\partial \hat{t}}\begin{pmatrix}P_\theta\\Q_\theta\end{pmatrix}/\delta_i+H(P_\theta,Q_\theta)\frac{\partial}{\partial z}\begin{pmatrix}P_\theta\\Q_\theta\end{pmatrix}-B(P_\theta,Q_\theta)\right)(s_i,1,(z_j,\hat{t}_j))\right|^2
\end{align*}
where $\{(z_j,\hat{t}_j)\}_{j=1}^M$ is randomly sampled from the domain $[0,L]\times[0,1] $. We approximate $\mathcal{R}_{pde}$ by
\begin{equation}
    \frac{1}{N}\sum_{i=1}^N\sum_{j=1}^M \left| \left(\frac{\partial}{\partial \hat{t}}\begin{pmatrix}P_\theta\\Q_\theta\end{pmatrix}/\delta_i+H(P_\theta,Q_\theta)\frac{\partial}{\partial z}\begin{pmatrix}P_\theta\\Q_\theta\end{pmatrix}-B(P_\theta,Q_\theta)\right)(s_i,1,(z_j,\hat{t}_j))\right|^2\label{eq:q_rpde}
\end{equation}
Then, the partial derivatives in (\ref{eq:q_rpde}) can be calculated as 
$$ \frac{\partial}{\partial x} P_\theta(s_i,\delta_i,(z_j,\hat{t}_j))=\mathbf{b_1}(s_i)^\top\frac{\partial}{\partial x} \mathbf{k}(1,(z_j,\hat{t}_j)),\quad x=z \text{ or } \hat{t}, $$
$$ \frac{\partial}{\partial x} Q_\theta(s_i,1,(z_j,\hat{t}_j))=\mathbf{b_2}(s_i)^\top\frac{\partial}{\partial x} \mathbf{k}(1,(z_j,\hat{t}_j)),\quad x=z \text{ or } \hat{t}. $$
For different $i$, we do not need to recompute the gradient of $\mathbf{k}$, so the number of gradient evaluations in one training epoch is only $O(M)$. We can use the same idea to approximate $\mathcal{R}_{b2}$ by
\begin{align}
    \mathcal{R}_{b2}(P_\theta,Q_\theta) 
    \approx \frac{1}{N}\sum_{i=1}^N\sum_{j=1}^M \left| \left(Q_\theta(1+R_1/R_2)+CR_1\frac{\partial}{\partial\hat{t}}Q_\theta-P_\theta/R_2-C\frac{\partial}{\partial\hat{t}}P_\theta\right)(s_i,1,(L,\hat{t}_j)).\right|^2\label{eq:q_rb2}
\end{align}
For $\mathcal{R}_{b1}$, we simply approximate it by
\begin{align}
    \mathcal{R}_{b1}(P_\theta) 
    \approx \frac{1}{N}\sum_{i=1}^N\sum_{j=1}^M \left|P_\theta(L,\hat{t}_j)-\bar{p}_i(\hat{t}_j\delta_i)\right|^2\label{eq:q_rb1}
\end{align}
where $\{\hat{t}_j\}_{j=1}^M$ is randomly sampled from the domain $[0,1]$ and $\bar{p}_i$ is the average of $p_i$ over all periods contained in $[0,T]$. Finally, the entire empirical loss function for training the BP-DeepONet is the combination of three terms (\ref{eq:q_rpde}), (\ref{eq:q_rb2}), and (\ref{eq:q_rb1}):
\begin{align*}
    \hat{\mathcal{L}}(P_\theta,Q_\theta)=& \frac{1}{N}\sum_{i=1}^N\sum_{j=1}^M \left| \left(\frac{\partial}{\partial \hat{t}}\begin{pmatrix}P_\theta\\Q_\theta\end{pmatrix}/\delta_i+H(P_\theta,Q_\theta)\frac{\partial}{\partial z}\begin{pmatrix}P_\theta\\Q_\theta\end{pmatrix}-B(P_\theta,Q_\theta)\right)(s_i,1,(z_j,\hat{t}_j))\right|^2\\
    &+\frac{\omega_1}{N}\sum_{i=1}^N\sum_{j=1}^M \left|P_\theta(s_i,1,(L,\hat{t}_j))-\bar{p}_i(\hat{t}_j\delta_i)\right|^2\\
    &+\frac{\omega_2}{N}\sum_{i=1}^N\sum_{j=1}^M \left| \left(Q_\theta(1+R_1/R_2)+CR_1\frac{\partial}{\partial\hat{t}}Q_\theta-P_\theta/R_2-C\frac{\partial}{\partial\hat{t}}P_\theta\right)(s_i,1,(L,\hat{t}_j))\right|^2.
\end{align*}

\subsection{Meta BP-DeepONet for sample-dependent hyper-parameters estimation}
In the Navier-Stokes equation, many hyper-parameters depend on the physical properties of arterial systems. In the previous BP-DeepONet, we set all these hyper-parameters to be subject-independent, i.e., we use the same set of hyper-parameters for all samples. However, in real situations, these parameters differ from person to person, so it is unrealistic to set them all as constants. To handle this issue, we generalize the model (\ref{eq:NS_system_sampleindependent}) to be sample-dependent. For each training sample $i$ from $1$ to $N$, we denote the set of all hyper-parameters for the $i$-th sample as
$$ \gamma_i:=\{\beta_i,A_{0,i},\rho_i,K_{r,i},P_{ext,i},R_{1,i},R_{2,i},C_i\}, $$ 
and we assume the solutions $P_i, Q_i$ for the $i$-th sample satisfy the following PDE: 

\begin{equation}
    \frac{\partial U_i}{\partial t}+H(U_i,\gamma_i)\frac{\partial U_i}{\partial z}=B(U_i,\gamma_i),\quad (z,t)\in[0,L]\times [0,T]
    \label{eq:NS_system_sampledependent}
\end{equation}
where
\begin{equation*}
    U_i=\begin{pmatrix}
        P_i \\ Q_i
    \end{pmatrix},
\end{equation*}
\begin{equation*}
    H(U_i,\gamma_i)=\begin{pmatrix}
        0 & \frac{\beta_i}{2a(P_i,\gamma_i)A_{0,i}} \\ \frac{a(P_i,\gamma_i)^2}{\rho_i}-\frac{2A_{0,i}Q_i^2}{\beta_i a^3(P_i,\gamma_i)} & \frac{2Q_i}{a(P_i,\gamma_i)^2}
    \end{pmatrix},
\end{equation*}
\begin{equation*}
    B(U_i,\gamma_i)=\begin{pmatrix}
        0 \\ -K_{r,i}\frac{Q_i}{a(P_i,\gamma_i)^2}
    \end{pmatrix},
\end{equation*}
and
\begin{equation*}
    a(P_i,\gamma_i)=\frac{A_{0,i}}{\beta_i}(P_i-P_{ext,i})+\sqrt{A_{0,i}}.
\end{equation*}
The corresponding boundary and periodic conditions are:
\begin{equation*}
    P_i(L,t)=p_i(t),\quad t\in[0,T],
\end{equation*}
\begin{equation}
    W(P_i,Q_i,\gamma_i):=Q_i(L,t)(1+\frac{R_{1,i}}{R_{2,i}})+C_iR_{1,i}\frac{\partial Q_i}{\partial t}(L,t)-\frac{P_i(L,t)}{R_{2,i}}-C_i\frac{\partial P_i}{\partial t}(L,t)=0,\quad t\in[0,T] \label{eq:Wind_bc_sampledependent},
\end{equation}
and
\begin{equation*}
    U_i(z,t+\delta_i)=U_i(z,t),\quad z\in[0,L].
\end{equation*}
It is almost impossible to measure $\gamma_i$ for each sample, so we adopt the idea from the field of meta-learning to estimate $\gamma_i$. We further define a hyper-network on top of the branch net. This network takes $\mathbf{b}_1$ and $\mathbf{b}_2$ as inputs and outputs the estimation of $\gamma_i$. This hyper-network is optimized together with the BP-DeepONet. We also impose extra constraints to regularize the output of the hyper-network. The first constraint is imposed as:
\begin{equation}
    \int_{0}^T P_i(z,t)dt = \int_{0}^T p_i(t)dt,\quad \forall z\in[0,L],
\end{equation}
which implies that the mean blood pressure does not change during the wave propagation. This phenomenon was found in clinical observations \cite{avolio2009role}. 
The second constraint is imposed as
\begin{equation}
    R_{1,i}+R_{2,i}=\frac{\int_{0}^{\delta_i} P_i(L,t)dt}{\int_{0}^{\delta_i} Q_i(L,t)dt}.
\end{equation}
This equation can be derived by integrating the equation (\ref{eq:Wind_bc_sampledependent}) over one period $[0,\delta_i]$ and it is used for regularizing the estimation of $R_{1,i}$ and $R_{2,i}$. If we have other information about the hemodynamics, like the pulse wave velocity, we may introduce them as constraints to regularize the hyper-networks. Then, the training loss function of the meta BP-DeepONet is 
\begin{align}
    \mathcal{L}_{meta}(P_\theta,Q_\theta,\gamma_\theta)=&\mathcal{R}_{pde}(P_\theta,Q_\theta,\gamma_\theta)\notag\\
    &+\omega_1\mathcal{R}_{b_1}(P_\theta)+\omega_2\mathcal{R}_{b_2}(P_\theta,Q_\theta,\gamma_\theta)+\omega_3\mathcal{R}_{c_1}(P_\theta,Q_\theta)+\omega_4\mathcal{R}_{c_2}(P_\theta,Q_\theta,\gamma_\theta) \label{eq:training_loss_meta}
\end{align}
where $\Gamma_\theta$ is the set of estimated hyper-parameters given by the hyper-network:
$$ \gamma_\theta(s_i)=\{\beta_\theta(s_i),A_{0,\theta}(s_i),\rho_\theta(s_i),K_{r,\theta}(s_i),P_{ext,\theta}(s_i),R_{1,\theta}(s_i),R_{2,\theta}(s_i),C_\theta(s_i)\},\quad i=1,\dots,N, $$
and the residual terms in (\ref{eq:training_loss_meta}) are defined as 
\begin{equation*}
    \mathcal{R}_{pde}(P_\theta,Q_\theta,\gamma_\theta)=\frac{1}{N}\sum_{i=1}^N\left\Vert \left(\frac{\partial}{\partial t}\begin{pmatrix}P_\theta\\Q_\theta\end{pmatrix}+H(P_\theta,Q_\theta,\gamma_\theta)\frac{\partial}{\partial z}\begin{pmatrix}P_\theta\\Q_\theta\end{pmatrix}-B(P_\theta,Q_\theta,\gamma_\theta)\right)(s_i,\delta_i,\cdot)\right\Vert^2_{L^2([0,L]\times[0,T])},
\end{equation*}
\begin{equation*}
    \mathcal{R}_{b_2}(P_\theta,Q_\theta,\gamma_\theta)=\frac{1}{N}\sum_{i=1}^N\left\Vert W(P_\theta,Q_\theta,\gamma_\theta)(s_i,\delta_i,(L,\cdot))\right\Vert_{L^2([0,T])}^2,
\end{equation*}
\begin{equation*}
    \mathcal{R}_{c_1}(P_\theta,Q_\theta)=\frac{1}{N}\sum_{i=1}^N\left\Vert \int_0^TP_\theta(s_i,\delta_i,(\cdot,t))dt-\int_0^Tp_i(t)dt\right\Vert_{L^2([0,L])}^2,
\end{equation*}
\begin{equation*}
    \mathcal{R}_{c_2}(P_\theta,Q_\theta,\gamma_\theta)=\frac{1}{N}\sum_{i=1}^N\left( R_{1,\theta}(s_i)+R_{2,\theta}(s_i)-\frac{\int_0^{\delta_i} P_\theta(s_i,\delta_i,(L,t))dt}{\int_0^{\delta_i} Q_\theta(s_i,\delta_i,(L,t))dt}\right)^2,
\end{equation*}
and $\mathcal{R}_{b_1}(P_\theta)$ is defined the same as (\ref{eq:training_loss}). To approximate these residual terms, we use the same change of variable technique as section \ref{sec:BP-DeepONet}:
\begin{equation*}
    \mathcal{R}_{pde}(P_\theta,Q_\theta,\gamma_\theta)\approx\frac{1}{N}\sum_{i=1}^N\sum_{j=1}^M \left|\left(\frac{\partial}{\partial t}\begin{pmatrix}P_\theta\\Q_\theta\end{pmatrix}+H(P_\theta,Q_\theta,\gamma_\theta)\frac{\partial}{\partial z}\begin{pmatrix}P_\theta\\Q_\theta\end{pmatrix}-B(P_\theta,Q_\theta,\gamma_\theta)\right)(s_i,1,(z_j,\hat{t}_j))\right|^2,
\end{equation*}
\begin{equation*}
    \mathcal{R}_{b_2}(P_\theta,Q_\theta,\gamma_\theta)\approx\frac{1}{N}\sum_{i=1}^N\sum_{j=1}^M\left| \left(Q_\theta(1+R_{1,\theta}/R_{2,\theta})+C_\theta R_{1,\theta}\frac{\partial}{\partial\hat{t}}Q_\theta-P_\theta/R_{2,\theta}-C_\theta\frac{\partial}{\partial\hat{t}}P_\theta\right)(s_i,1,(L,\hat{t}_j))\right|^2,
\end{equation*}
\begin{equation*}
    \mathcal{R}_{c_1}(P_\theta,Q_\theta)\approx\frac{1}{N}\sum_{i=1}^N\sum_{j_1=1}^{M'}\left|\sum_{j_2=1}^M (P_\theta(s_i,1,(\hat{z}_{j_1},\hat{t}_{j_2}))-p_i(\hat{t}_{j_2}))\right|^2,
\end{equation*}
\begin{equation*}
    \mathcal{R}_{c_2}(P_\theta,Q_\theta,\gamma_\theta)\approx\frac{1}{N}\sum_{i=1}^N\left| R_{1,\theta}(s_i)+R_{2,\theta}(s_i)-\frac{\sum_{j=1}^M P_\theta(s_i,\delta_i,(L,\hat{t}_j))}{\sum_{j=1}^M Q_\theta(s_i,\delta_i,(L,\hat{t}_j))}\right|^2,
\end{equation*}
where $\{(z_j,\hat{t}_j)\}_{j=1}^M$ is a set of quadrature points sampled from $[0,L]\times[0,1]$ and $\{\hat{z}_j\}_{j=1}^{M'}$ is a set of quadrature points sampled from $[0,L]$,
and $\mathcal{R}_{b_1}(P_\theta)$ is approximated in the same way as (\ref{eq:q_rb1}). Then, the empirical training loss for the meta BP-DeepONet is given as 
\begin{align*}
    \hat{\mathcal{L}}_{meta}(P_\theta,Q_\theta,\gamma_\theta)= & \frac{1}{N}\sum_{i=1}^N\sum_{j=1}^M \left|\left(\frac{\partial}{\partial t}\begin{pmatrix}P_\theta\\Q_\theta\end{pmatrix}+H(P_\theta,Q_\theta,\gamma_\theta)\frac{\partial}{\partial z}\begin{pmatrix}P_\theta\\Q_\theta\end{pmatrix}-B(P_\theta,Q_\theta,\gamma_\theta)\right)(s_i,1,(z_j,\hat{t}_j))\right|^2\\
    &+\frac{\omega_1}{N}\sum_{i=1}^N\sum_{j=1}^M \left|P_\theta(s_i,1,(L,\hat{t}_j))-\bar{p}_i(\hat{t}_j\delta_i)\right|^2\\
    &+\frac{\omega_2}{N}\sum_{i=1}^N\sum_{j=1}^M\left| \left(Q_\theta(1+R_{1,\theta}/R_{2,\theta})+C_\theta R_{1,\theta}\frac{\partial}{\partial\hat{t}}Q_\theta-P_\theta/R_{2,\theta}-C_\theta\frac{\partial}{\partial\hat{t}}P_\theta\right)(s_i,1,(L,\hat{t}_j))\right|^2\\
    &+\frac{\omega_3}{N}\sum_{i=1}^N\sum_{j_1=1}^{M'}\left|\sum_{j_2=1}^M (P_\theta(s_i,1,(\hat{z}_{j_1},\hat{t}_{j_2}))-p_i(\hat{t}_{j_2}))\right|^2\\
    &+\frac{\omega_4}{N}\sum_{i=1}^N\left| R_{1,\theta}(s_i)+R_{2,\theta}(s_i)-\frac{\sum_{j=1}^M P_\theta(s_i,\delta_i,(L,\hat{t}_j))}{\sum_{j=1}^M Q_\theta(s_i,\delta_i,(L,\hat{t}_j))}\right|^2.
\end{align*}

\begin{figure}
    \centering
    \includegraphics[width=1\textwidth]{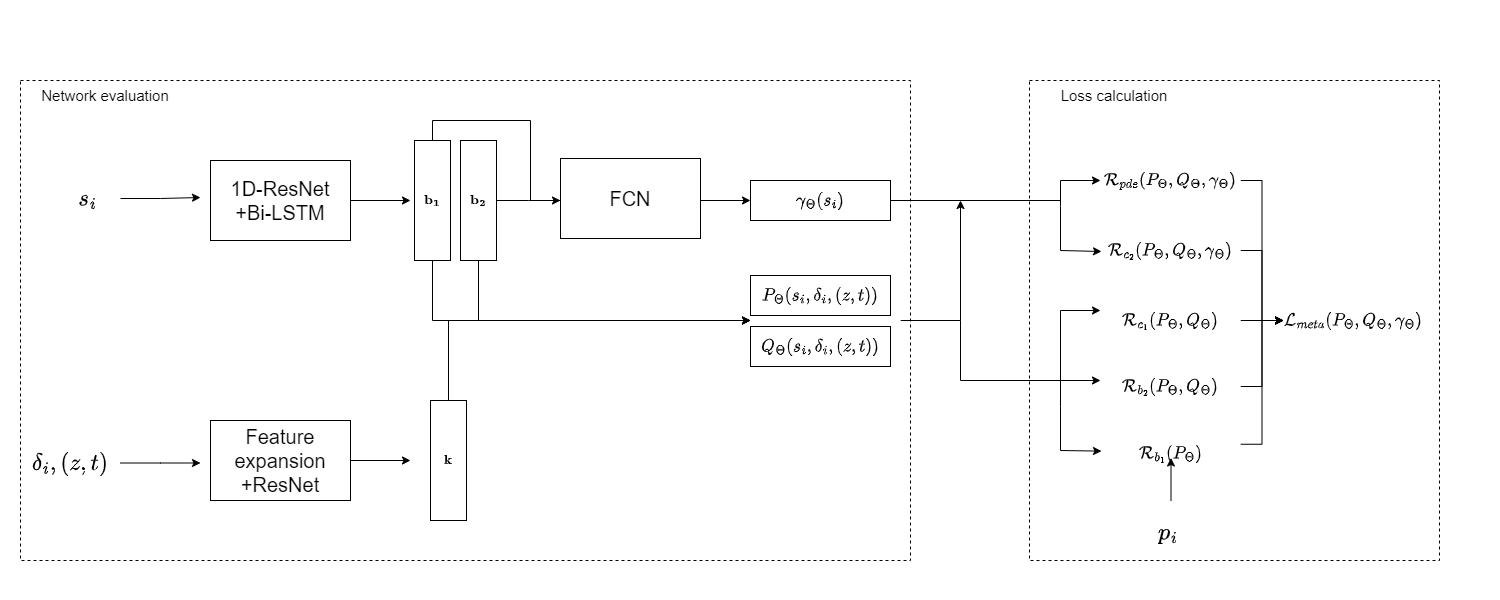}
    \caption{Strucutres of the meta BP-DeepONet}
    \label{fig:meta_bp_deeponet}
\end{figure}

The overall structures of the meta BP-DeepONet are given in Figure \ref{fig:meta_bp_deeponet} where the hyper-network is implemented as a simple FCN.

\begin{remark}
    During the training of the meta BP-DeepONet, we only require the measurement of blood pressure at $z=L$. Then the whole solution, all boundary conditions, initial conditions, and hyper-parameters can be automatically learned from this measurement. No extra measurements and parameters are needed. 
\end{remark}

\section{Numerical experiments of the PINN on simulated data}
In this section, we would like to validate the proposed physics-informed training method by solving one PDE instance. We first generate a solution to the Navier-Stokes equation (\ref{eq:NS_system}) using a MacCormack scheme with a prescribed initial condition and inlet boundary condition. Then, we solve this equation with the boundary conditions (\ref{eq:Dirchlet_bc}) and (\ref{eq:Wind_bc}) and the time-periodic condition (\ref{eq:periodic}) using a simple PINN. We can see in the numerical results that the PINN solution is very close to the simulated solution.

\begin{figure}
    \centering
    \includegraphics[width=0.4\textwidth]{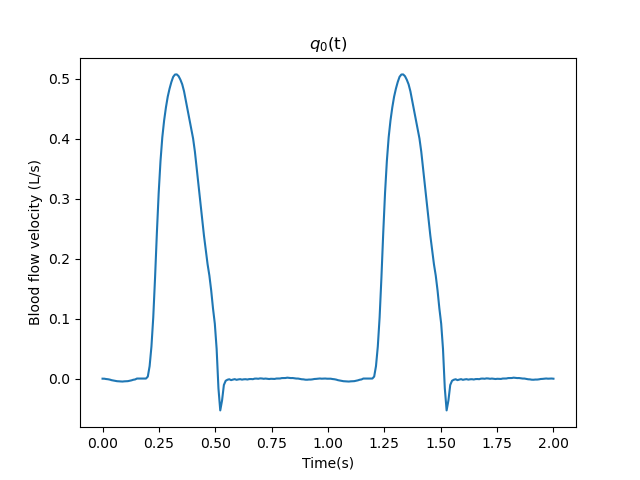}
    \caption{The inflow boundary condition}
    \label{fig:inflow}
\end{figure}

We consider the 1D Navier-stokes equation (\ref{eq:NS_1}) and (\ref{eq:NS_2}) with the initial condition:
$$ A(x,0)=A_0,\quad Q(x,0)=0, $$
the inflow boundary condition:
$$ Q(0,t)=q_0(t), $$
and the three-element Windkessel boundary condition
$$ Q(L,t)(1+\frac{R_{1}}{R_{2}})+CR_{1}\frac{\partial Q}{\partial t}(L,t)=\frac{P(L,t)}{R_{2}}+C\frac{\partial P}{\partial t}(L,t). $$
Such an initial boundary value problem can be efficiently solved by many classical methods like the MacCormack scheme\cite{elad1991numerical,fullana2009branched}, Taylor-Galerkin scheme \cite{wan2002one,sherwin2003computational}, MUSCL \cite{cavallini2008finite,delestre2013well}, and the local discontinuous Galerkin scheme \cite{matthys2007pulse,marchandise2009numerical,mynard20081d}. We use a numerical example that is similar to the baseline aorta case provided in \cite{xiao2014systematic}, where the hyper-parameters are chosen as $L=25\text{cm}$, $A_0=4.52\text{cm}^2$, $\beta=1134.37 \text{kg/s}^2$, $\rho=1060\text{kg/m}^3$, $P_{ext}=9\text{kPa}$, $R_1=1.17\times 10^7\text{Pa s m}^{-3}$, $R_1=1.12\times 10^8\text{Pa s m}^{-3}$, $C=1.01\times 10^{-8}\text{m}^3\text{Pa}^{-1}$, $v=0.01\times 10^6\text{m}^2/\text{s}^2$, and the inflow boundary conditions $q_0(t)$ is given in Figure \ref{fig:inflow}. Then, we solve this problem using the MacCormack scheme. 

\begin{figure}
    \centering
    \includegraphics[width=\textwidth]{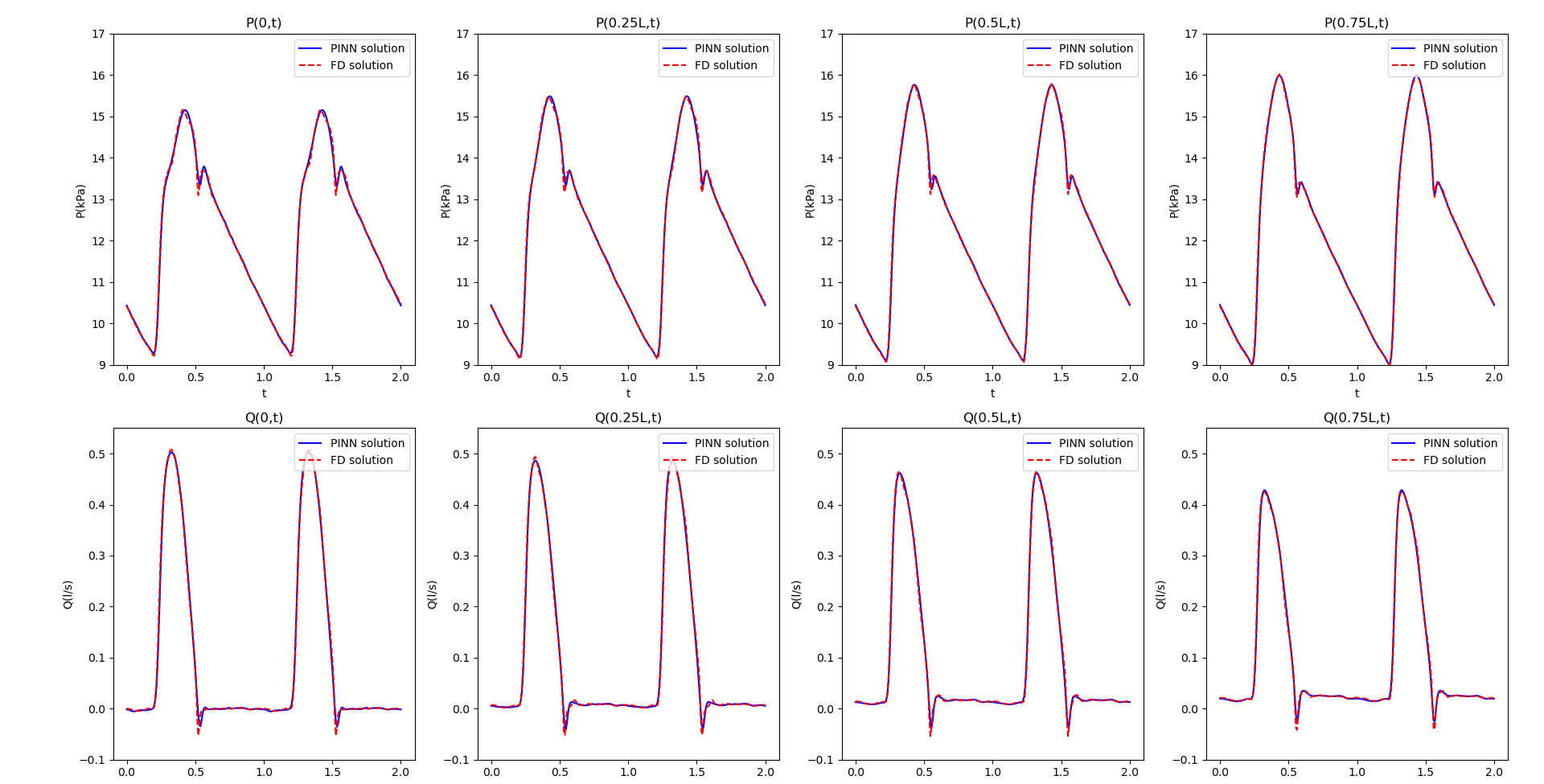}
    \caption{Comparison of the PINN solution and the finite difference solution (MacCormack).}
    \label{fig:PINN-FD}
\end{figure}

After having a referenced numerical solution $\hat{A}$ and $\hat{Q}$, we define a residual network $N_\theta(z,t)=(P_\theta(z,t),Q_\theta(z,t))$ shown in Figure \ref{fig:structure1} (b) with five residual layers. The time-periodic condition is incorporated into the network using the same feature expansion technique described before. We train this network by minimizing the following loss function:
\begin{align*}
    L_{PINN}&(\theta):=\left\Vert \frac{\partial A_\theta}{\partial t}+\frac{\partial Q_\theta}{\partial z}\right\Vert_{L^2([0,L]\times[0,T])}^2+\left\Vert \frac{\partial Q_\theta}{\partial t}+\frac{\partial}{\partial z}(\frac{Q_\theta^2}{A_\theta})+\frac{A_\theta}{\rho}\frac{\partial P_\theta}{\partial z}+K_r\frac{Q_\theta}{A_\theta}\right\Vert_{L^2([0,L]\times[0,T])}^2\\
    &+100\left(\Vert P_\theta(L,\cdot)-\hat{P}(L,\cdot)\Vert_{L^2([0,T])}^2+\left\Vert Q_\theta(L,\cdot)(1+\frac{R_{1}}{R_{2}})+CR_{1}\frac{\partial Q_\theta}{\partial t}(L,\cdot)-\frac{P_\theta(L,\cdot)}{R_{2}}-C\frac{\partial P_\theta}{\partial t}(L,\cdot)\right\Vert_{L^2([0,T])}^2\right)
\end{align*}
where $P_\theta(z,t)=P_{ext}+\frac{\beta}{A_0}(\sqrt{A_\theta(z,t)}-\sqrt{A_0})$. We use a standard Adam optimizer with a learning rate of 1e-4, and the total number of training epochs is 50,000. We randomly sample 1,000 quadrature points in the domain in each epoch to approximate the $L^2$ integration. We can see the comparison of the PINN solution and the numerical solution by the MacCormack scheme in Figure \ref{fig:PINN-FD}. The PINN solution solved using the loss function $L_{PINN}$ is very close to the solution of the MacCormack scheme. 
We also calculate the relative error:
$$ \left\Vert\frac{\hat{P}-P_\theta}{\hat{P}}\right\Vert_1\approx 0.30\%$$
and
$$ \frac{\Vert \hat{Q}-Q_\theta\Vert_1}{\Vert \hat{Q}\Vert_{\infty}} \approx 0.67\%.$$
These two metrics are also used in \cite{xiao2014systematic} to measure the error of their numerical solutions for the Navier-Stokes equation. From Figure \ref{fig:PINN-FD}, we can observe that the error of the PINN solution is larger near the dicrotic notch where the solution is near singular. This case is very common when using the standard PINN method to solve PDEs, and it can be improved by using some adaptive sampling scheme \cite{gao2023failure}. We may consider this direction in our future work.

\section{Numerical experiments of the PI-DeepONet methods on clinical data}
\subsection{Dataset description}
We conducted our experiments on the MIMIC dataset \cite{johnson2016mimic}, the largest and most commonly used dataset for cuffless blood pressure estimation. We use the pre-processed version of the MIMIC data from \cite{misc_cuff-less_blood_pressure_estimation_340}, which contains recordings of 12,000 subjects after simple pre-processing. Each recording consists of continuous ECG, PPG, and ABP waveform data. The length of each recording is within 10 minutes. All these data are collected from ICU patients, and the invasive method collects the ABP waveform data.
Furthermore, we also apply some extra pre-processing steps to these data:
\begin{enumerate}
    \item Use a bandpass filter to remove noise from PPG and ABP signals.
    \item Align the PPG and ABP signals.
    \item Randomly select 75\% of subjects for training and 25\% for testing.
    \item 20\% of the training set is selected as the validation set.
    \item The first 15\% of data in each testing subject is used for calibration.
    \item Split each recording into many 4-second segments without overlapping.
    \item Remove those segments that are highly non-periodic.
    \item Remove subjects whose number of valid segments is less than 10.
    \item Apply some peak detection algorithms to detect all peaks in each PPG segment, then calculate the average length of intervals between two consecutive peaks.
\end{enumerate}

\begin{figure}[h]
    \centering
    \subfigure[]{\includegraphics[width=0.3\textwidth]{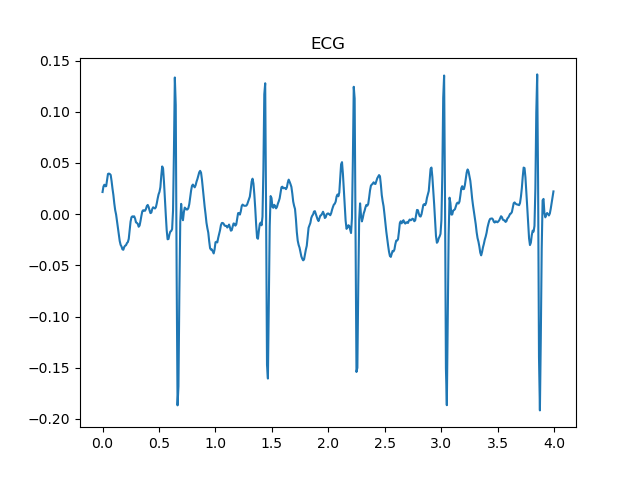}} 
    \subfigure[]{\includegraphics[width=0.3\textwidth]{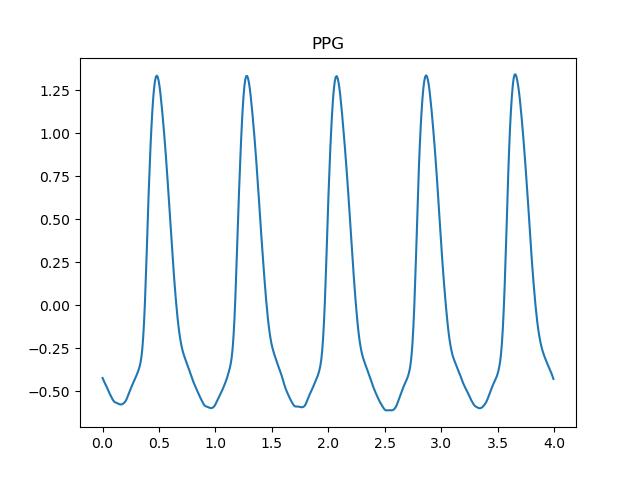}} 
    \subfigure[]{\includegraphics[width=0.3\textwidth]{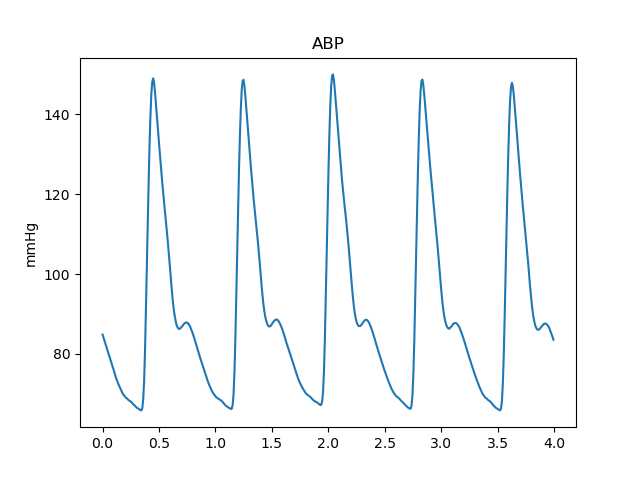}} 
    \caption{One segment of ECG, PPG, and ABP signals.}
    \label{fig:ecg_ppg_abp}
\end{figure}

After pre-processing, the total number of training and testing samples is about 70,000. Step 5 is also called individual fine-tuning \cite{hong2021deep}, and it is a necessary process to calibrate and improve the blood pressure measurement. The segments for calibration will not used when calculating the testing error. One segment of the recording is shown in Figure \ref{fig:ecg_ppg_abp}. 

\subsection{Training procedures}
For the training of the BP-DeepONet and the Meta BP-DeepONet, We apply the standard Adam optimizer with a scheduled learning rate to train the network. The initial learning rate is set to be 1e-4, and decays by a factor of 0.9 every 5 epochs. After 200 epochs, we restart the learning rate with 1e-4 and continue to train with the same learning rate schedule. After several rounds of learning rate cycles, we choose the best model based on the performance evaluated on a validation set.

When training the BP-DeepONet, we need first to determine all the hyper-parameters in the NS model, so we set hyper-parameters based on the reference values of the radial artery described in \cite{charlton2019modeling} and estimate the Windkessel parameters using the method provided in \cite{xiao2014systematic}: $A_0=8\times 10^{-6}\ m^2$, $\beta=529\ kg/s^2$, $L=0.25m$, $K_r=8*\pi*0.01\approx 0.251$, $\rho=1.06\ kg/m^3$, $P_{ext}=6.67\ kPa$, $R_1=12.37\ Pa\cdot s/m^3$, $R_2=137.5\ Pa\cdot s/m^3$, $C=0.01\ m^3/Pa$. In the Meta BP-DeepONet, we also use these values as referenced values for the predicted hyper-parameters. We restrict the deviation of the predicted hyper-parameters from the referenced values to be less than or equal to 20 \% so that the estimation of the hyper-parameters will always be reasonable. 

During the training of both BP-DeepONet and meta BP-DeepONet, we use a batch size of $100$. All the training is done on a GPU server with 4 NVIDIA RTX A6000 GPUs.

\subsection{Numerical results}
After finishing the training of DeepONet, we can obtain the prediction of the BP waveform at any given location $z$ by evaluating the DeepONet at $(z,t_0)$, $(z,t_1)$, \dots, $(z,t_{N_h})$, where $\{t_j\}_{j=0}^{N_h}$ is a uniform discretization of the domain $[0,T]$. 

We first evaluate the waveform prediction at the outlet side $z=L$. Since we have the measured BP wave at $z=L$ by the invasive method, we can compute the prediction error of our method at this location. We evaluate the error using several metrics: waveform MAE, SBP MAE, DBP MAE, and MBP MAE, where MAE is short for mean absolute error. Suppose $P_{i}=(P_{\theta}(s_i,\delta_i,(L,t_0)), P_{\theta}(s_i,\delta_i,(L,t_1)),\dots,P_{\theta}(s_i,\delta_i,(L,t_{N_h})))\in\mathbb{R}^{N_h}$ is the predicted waveform and $p_i\in\mathbb{R}^{N_h}$ is the ground true BP waveform. Then, the four metrics are defined as follows:
\begin{itemize}
    \item Waveform MAE: $\frac{1}{N}\frac{1}{N_h}\sum_{i=1}^N|P_i-p_i|_1$,
    \item SBP MAE: $\frac{1}{N}\sum_{i=1}^N |\max(P_i)-\max(p_i)|$,
    \item DBP MAE: $\frac{1}{N}\sum_{i=1}^N |\min(P_i)-\min(p_i)|$,
    \item MBP MAE: $\frac{1}{N}\sum_{i=1}^N |\text{mean}(P_i)-\text{mean}(p_i)|$,
\end{itemize}
where $|\cdot|_1$ denotes the standard vector $l_1$ norm. 
In Table \ref{table:prediction_error}, we listed the error metrics for both BP-DeepONet and meta BP-DeepONet. We also compared our results with the IEEE standard for wearable, cuffless blood pressure measuring devices \cite{ieee2014ieee}. The suggested grading is also shown in Table \ref{table:prediction_error}. We can see the meta BP-DeepONet achieves better accuracy on the MAE of SBP, DBP, and MBP, while the BP-DeepONet achieves better accuracy on the waveform MAE. According to the IEEE standard, the meta BP-DeepONet is graded as A-level in all metrics, while the BP-DeepONet is graded as A-level except on the SBP measurement.

\begin{table}[h]
\centering
\begin{tabular}{|p{0.2\textwidth}|p{0.2\textwidth}|p{0.1\textwidth}p{0.1\textwidth}p{0.1\textwidth}p{0.1\textwidth}|}
\hline
      & Methods & Waveform &SBP  &DBP  & MBP  \\ \hline
\multirow{2}{*}{MAE (mmHg)} & BP-DeepONet          & \textbf{4.406}               & 5.057                       & 2.581                       & 2.881                       \\
                           & Meta BP-DeepONet     & 4.555                        & \textbf{4.934}              & \textbf{2.574}              & \textbf{2.871}              \\ \hline
\multirow{2}{*}{Grading}    & BP-DeepONet          & A                            & B                           & A                           & A                           \\
                            & Meta BP-DeepONet     & A                            & A                           & A                           & A                           \\ \hline
\end{tabular}
\caption{The prediction error of two proposed methods and their grading according to the IEEE standard.}
\label{table:prediction_error}
\end{table}

Since our methods can predict the blood pressure $P(z,t)$ and blood flow rate $Q(z,t)$ in the whole domain $(z,t)\in[0,L]\times[0,T]$, we also want to evaluate the errors in the interior of this domain. We choose two different metrics to evaluate the error in $(z,t)\in[0,L]\times[0,T]$. The first metric is the residual errors $\mathcal{R}_{pde}$. The residual error is usually used to estimate the total error when the true solution is unknown. For the BP-DeepONet, the residual error is defined as:
\begin{align*}
    & RE_1(P_\theta,Q_\theta)=\frac{1}{N}\sum_{i=1}^N \left\Vert \left(\frac{\partial A_\theta}{\partial t}+\frac{\partial Q_\theta}{\partial z}\right)(s_i,\delta_i,(\cdot,\cdot)) \right\Vert_{L^2([0,L]\times[0,T])}\\
    & RE_2(P_\theta,Q_\theta)=\frac{1}{N}\sum_{i=1}^N \left\Vert\left(\frac{\partial Q_\theta}{\partial t} + \frac{\partial}{\partial z}(\frac{Q_\theta^2}{A_\theta})+\frac{A_\theta}{\rho}\frac{\partial P_\theta}{\partial z}+K_r\frac{Q_\theta}{A_\theta}\right)(s_i,\delta_i,(\cdot,\cdot)) \right\Vert_{L^2([0,L]\times[0,T])}
\end{align*}
where $A_\theta=\left(\frac{A_0}{\beta}(P_\theta-P_{ext})+\sqrt{A_0}\right)^2$. For the meta BP-DeepONet, the residual error is defined as:
\begin{align*}
    & RE_1(P_\theta,Q_\theta,\gamma_\theta)=\frac{1}{N}\sum_{i=1}^N \left\Vert \left(\frac{\partial A_{\theta,i}}{\partial t}+\frac{\partial Q_\theta}{\partial z} \right)(s_i,\delta_i,(\cdot,\cdot))\right\Vert_{L^2([0,L]\times[0,T])}\\
    & RE_2(P_\theta,Q_\theta,\gamma_\theta)=\frac{1}{N}\sum_{i=1}^N \left\Vert \left(\frac{\partial Q_\theta}{\partial t} + \frac{\partial}{\partial z}(\frac{Q_\theta^2}{A_{\theta,i}})+\frac{A_{\theta,i}}{\rho_i}\frac{\partial P_\theta}{\partial z}+K_{r_i}\frac{Q_\theta}{A_{\theta,i}}\right)(s_i,\delta_i,(\cdot,\cdot)) \right\Vert_{L^2([0,L]\times[0,T])}
\end{align*}
where $A_{\theta,i}=\left(\frac{A_{0,i}}{\beta}(P_\theta-P_{ext,i})+\sqrt{A_{0,i}}\right)^2$. In \cite{mishra2022estimates}, the authors show that the error of solutions to the Navier-Stokes by neural networks can be bounded by the residual errors. The residual errors of the two methods are shown in Table \ref{table:res_error}. 

\begin{table}[h]
\centering
\begin{tabular}{|p{0.3\textwidth}|p{0.2\textwidth}p{0.2\textwidth}|}
\hline
       & $RE_1(\sqrt{L})$ & $RE_2(\sqrt{dm*L/s})$  \\ \hline
BP-DeepONet          & 1.11$\times$ 1e-4               & 2.65$\times$ 1e-2                   \\
Meta BP-DeepONet     & \textbf{0.22$\times$ 1e-4}                     & \textbf{1.14$\times$1e-2}              \\ \hline
\end{tabular}
\caption{The residual error of two proposed methods.}
\label{table:res_error}
\end{table}

The second metric we used to evaluate the error is the relative error compared to a fine-tuned BP-DeepONet (or meta BP-DeepONet) on each test sample. Though we do not know the ground true solution, we have shown in Section 5 that a well-trained PINN solution is reasonably accurate compared to the simulated true solution. Therefore, we can fine-tune the network on each test sample, i.e., minimize the PDE residual loss of this particular sample only, and use this fine-tuned network as an approximation to the true solution. During the fine-tuning stage, we fix the parameters in the branch net and only train the parameters in the trunk net so that predictions of hyper-parameters will not change. The error metric is defined as:
$$ \sum_{i=1}^N\frac{\left\Vert\left(\hat{P}_{tuned}-P_\theta\right)(s_i,\delta_i,(\cdot,\cdot))\right\Vert_{L^1([0,L]\times[0,T])}}{\left\Vert\hat{P}_{tuned}(s_i,\delta_i,(\cdot,\cdot))\right\Vert_{L^1([0,L]\times[0,T])}}$$
and
$$ \sum_{i=1}^N\frac{\left\Vert\left(\hat{Q}_{tuned}-Q_\theta\right)(s_i,\delta_i,(\cdot,\cdot))\right\Vert_{L^1([0,L]\times[0,T])}}{\left\Vert \hat{Q}_{tuned}(s_i,\delta_i,(\cdot,\cdot))\right\Vert_{L^\infty([0,L]\times[0,T])}} $$
where $\hat{P}_{tuned}$ and $\hat{Q}_{tuned}$ denote the solution after fine-tuning. The relative errors are given in Table \ref{table:relative_error}. We can see the BP-DeepONet can predict the PDE solution better than the Meta BP-DeepONet. This result is not surprising because the BP-DeepONet assumes all samples satisfy the same PDE, which greatly simplifies the problem. 

\begin{table}[h]
\centering
\begin{tabular}{|p{0.3\textwidth}|p{0.2\textwidth}p{0.2\textwidth}|}
\hline
       & error of $P_\theta$ & error of $Q_\theta$  \\ \hline
BP-DeepONet          & 4.10$\times$ 1e-2               & 5.35$\times$ 1e-2                   \\
Meta BP-DeepONet     & 4.18$\times$ 1e-2                    & 1.03$\times$1e-1              \\ \hline
\end{tabular}
\caption{The relative error of two proposed methods.}
\label{table:relative_error}
\end{table}

\begin{figure}
    \centering
    \subfigure[]{\includegraphics[width=0.3\textwidth]{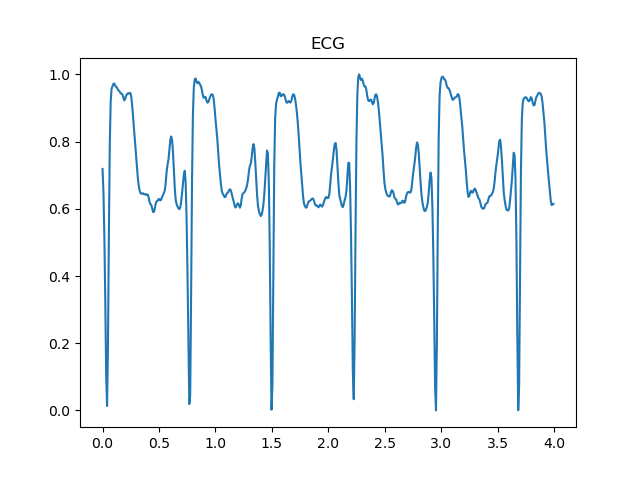}} 
    \subfigure[]{\includegraphics[width=0.3\textwidth]{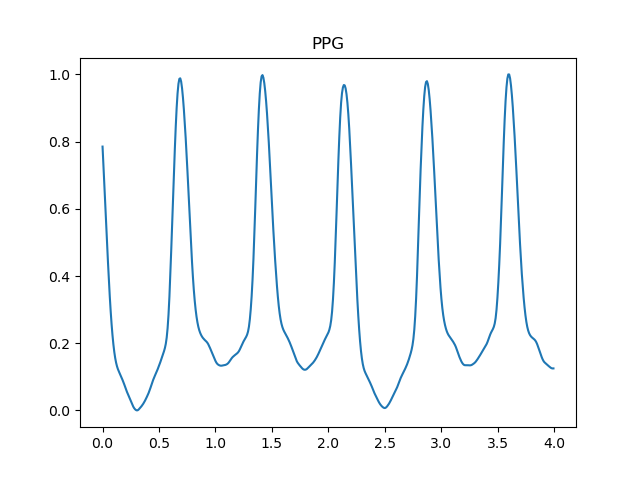}} 
    \subfigure[]{\includegraphics[width=0.3\textwidth]{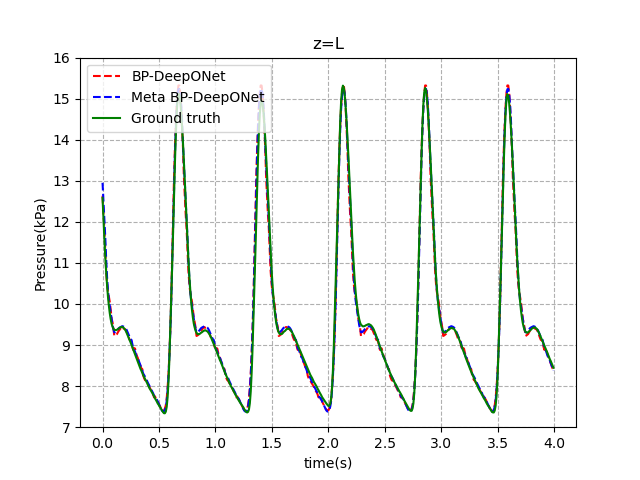}} 
    \caption{(a) and (b) show the input physiological signals. (c) shows the prediction and ground truth data at $z=L$.}
    \label{fig:bp_predict}
\end{figure}

\begin{figure}
    \centering
    \subfigure[]{\includegraphics[width=0.4\textwidth]{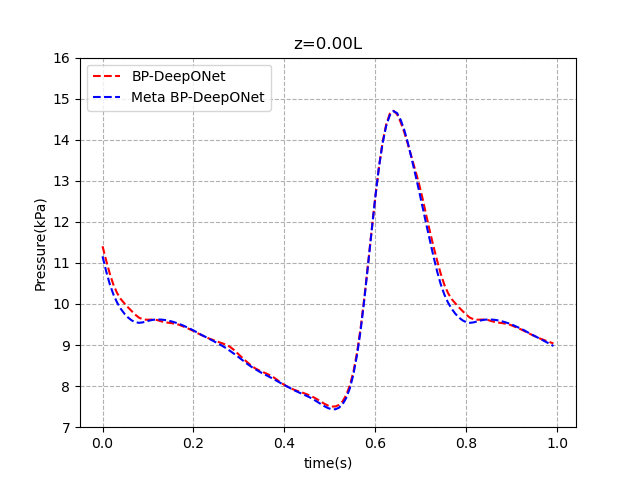}} 
    \subfigure[]{\includegraphics[width=0.4\textwidth]{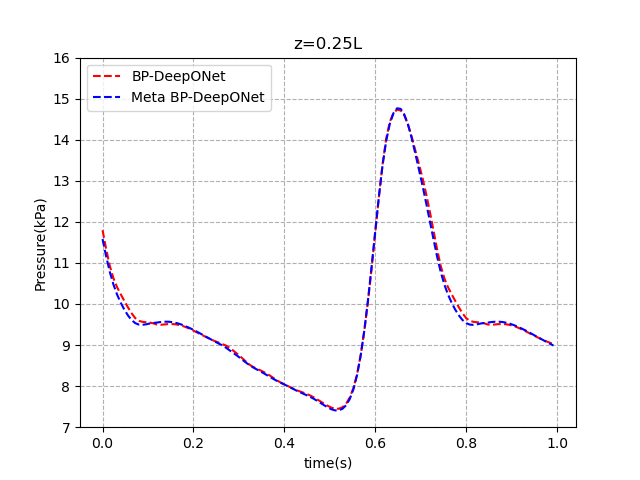}} 
    \subfigure[]{\includegraphics[width=0.4\textwidth]{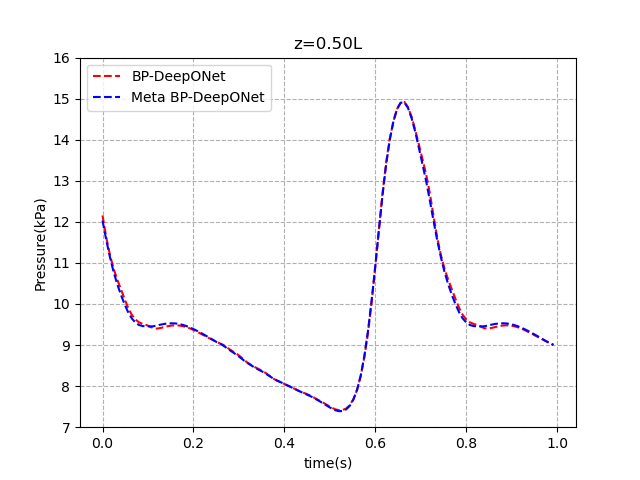}} 
    \subfigure[]{\includegraphics[width=0.4\textwidth]{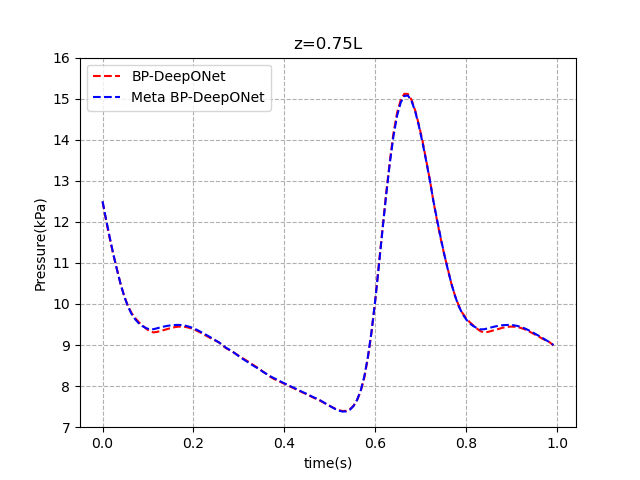}} 
    \caption{Predictions of ABP waveforms at different locations $z$.}
    \label{fig:bp_predict2}
\end{figure}

We also visualize the predicted waveforms for one test sample in Figure \ref{fig:bp_predict} and Figure \ref{fig:bp_predict2}. We can observe that the predictions of both the BP-DeepONet and the meta BP-DeepONet are very close to the ground truth data at the outlet side $z=L$ (Figure \ref{fig:bp_predict} (c)). We can also observe the pulse pressure amplification with the difference between the maximum and minimum of the waveforms increases when $z$ increases for both methods in Figure \ref{fig:bp_predict2}. Because the Meta BP-DeepONet uses different hyper-parameters for each sample, its prediction differs from BP-DeepONet in $z\in(0,1)$, especially at $z=0$. 

\section{Conclusions and Future Works}
In this work, we propose two methods based on the physics-informed DeepONet to predict the blood pressure waveform: the BP-DeepONet and the meta BP-DeepONet. We incorporate the Navier-Stokes equation with time-periodic conditions and Windkessel boundary conditions into the physics-informed training procedure and provide efficient implementation algorithms. The proposed methods are the first ones that can predict waveforms at different locations. During our numerical experiments, the proposed methods can predict blood pressure waveforms at reasonable accuracy and preserve some physical properties consistent with the clinical observations. 

This work only considers very standard DNN models for trunk and branch nets. We may try implementing different DNN architectures in the future, like the transformer, which is very powerful in processing time-series data. Besides, the inputs to the DeepONet are just the pre-processed physiological signals. We may also try including some morphological features in the inputs, effectively improving the prediction accuracy.

We know the arterial system of humans is very complicated. As the first work in this direction, we only consider a simple cylindrical segment of an artery. Still, we may generalize our methods to more complex geometry afterward—for example, a simple arterial tree with bifurcations.

In the future, we will also consider generating a simulated dataset to validate better the accuracy of both BP-DeepONet and meta BP-DeepONet methods.

\section*{Acknowledgement}
The work of Lingfeng Li is supported by the InnoHK project at Hong Kong Centre for Cerebro-cardiovascular Health Engineering (COCHE).
The work of Xue-Cheng Tai is partially supported by HKRGC-NSFC Grant N-CityU214/19, HKRGC CRF.
Grant C1013-21GF and NORCE Kompetanseoppbygging program.
The work of Raymond Chan is partially supported by HKRGC GRF
grants CityU1101120, CityU11309922, CRF grant C1013-21GF and HKITF MHKJFS Grant MHP/054/22.
We want to thank Prof Yuan-Ting Zhang and Dr. Alireza Keramat for the helpful discussion during this project.

\bibliographystyle{elsarticle-num}
\bibliography{references}

\begin{thebibliography}{10}
\expandafter\ifx\csname url\endcsname\relax
  \def\url#1{\texttt{#1}}\fi
\expandafter\ifx\csname urlprefix\endcsname\relax\def\urlprefix{URL }\fi
\expandafter\ifx\csname href\endcsname\relax
  \def\href#1#2{#2} \def\path#1{#1}\fi

\bibitem{WHO}
{World Health Organization}, Cardiovascular diseases (cvds),
  \url{https://www.who.int/news-room/fact-sheets/detail/cardiovascular-diseases-(cvds)}.

\bibitem{fuchs2020high}
F.~D. Fuchs, P.~K. Whelton, High blood pressure and cardiovascular disease,
  Hypertension 75~(2) (2020) 285--292.

\bibitem{matthieu2008cardiac}
B.~Matthieu, N.-G. Karine, C.~Vincent, V.~Alain, C.~J. Fran{\c{c}}ois,
  R.~Philippe, S.~Fran{\c{c}}ois, Cardiac output measurement in patients
  undergoing liver transplantation: pulmonary artery catheter versus
  uncalibrated arterial pressure waveform analysis, Anesthesia \& Analgesia
  106~(5) (2008) 1480--1486.

\bibitem{thiele2011arterial}
R.~H. Thiele, M.~E. Durieux, Arterial waveform analysis for the
  anesthesiologist: past, present, and future concepts, Anesthesia \& Analgesia
  113~(4) (2011) 766--776.

\bibitem{li2017forward}
Y.~Li, H.~Gu, H.~Fok, J.~Alastruey, P.~Chowienczyk, Forward and backward
  pressure waveform morphology in hypertension, Hypertension 69~(2) (2017)
  375--381.

\bibitem{hullender2021simulations}
D.~A. Hullender, O.~R. Brown, Simulations of blood pressure and identification
  of atrial fibrillation and arterial stiffness using an extended kalman filter
  with oscillometric pulsation measurements, Computer Methods and Programs in
  Biomedicine 198 (2021) 105768.

\bibitem{su2018long}
P.~Su, X.-R. Ding, Y.-T. Zhang, J.~Liu, F.~Miao, N.~Zhao, Long-term blood
  pressure prediction with deep recurrent neural networks, in: 2018 IEEE EMBS
  International Conference on Biomedical \& Health Informatics (BHI), IEEE,
  2018, pp. 323--328.

\bibitem{eom2020end}
H.~Eom, D.~Lee, S.~Han, Y.~S. Hariyani, Y.~Lim, I.~Sohn, K.~Park, C.~Park,
  End-to-end deep learning architecture for continuous blood pressure
  estimation using attention mechanism, Sensors 20~(8) (2020) 2338.

\bibitem{ibtehaz2022ppg2abp}
N.~Ibtehaz, S.~Mahmud, M.~E. Chowdhury, A.~Khandakar, M.~Salman~Khan, M.~A.
  Ayari, A.~M. Tahir, M.~S. Rahman, Ppg2abp: Translating photoplethysmogram
  (ppg) signals to arterial blood pressure (abp) waveforms, Bioengineering
  9~(11) (2022) 692.

\bibitem{ma2022kd}
C.~Ma, P.~Zhang, F.~Song, Y.~Sun, G.~Fan, T.~Zhang, Y.~Feng, G.~Zhang,
  Kd-informer: Cuff-less continuous blood pressure waveform estimation approach
  based on single photoplethysmography, IEEE Journal of Biomedical and Health
  Informatics (2022).

\bibitem{lakhal2017invasive}
K.~Lakhal, V.~Robert-Edan, Invasive monitoring of blood pressure: a radiant
  future for brachial artery as an alternative to radial artery
  catheterisation?, Journal of Thoracic Disease 9~(12) (2017) 4812.

\bibitem{2008ContinuousNA}
S.-P. E, \href{https://api.semanticscholar.org/CorpusID:30046018}{Continuous
  non-invasive arterial pressure shows high accuracy in comparison to invasive
  intra-arterial blood pressure measurement}, 2008.
\newline\urlprefix\url{https://api.semanticscholar.org/CorpusID:30046018}

\bibitem{armstrong2019brachial}
M.~K. Armstrong, M.~G. Schultz, D.~S. Picone, J.~A. Black, N.~Dwyer,
  P.~Roberts-Thomson, J.~E. Sharman, Brachial and radial systolic blood
  pressure are not the same: evidence to support the popeye phenomenon,
  Hypertension 73~(5) (2019) 1036--1041.

\bibitem{kyriazis2001dp}
J.~Kyriazis, J.~Glotsos, L.~Bilirakis, N.~Smirnioudis, The (dp/dt) max derived
  from arterial pulse waveforms: prospective applications in the haemodialysis
  setting, Nephrology Dialysis Transplantation 16~(5) (2001) 1087--1088.

\bibitem{sharman2007radial}
J.~Sharman, A.~Qasem, L.~Hanekom, D.~Gill, R.~Lim, T.~Marwick, Radial pressure
  waveform dp/dt max is a poor indicator of left ventricular systolic function,
  European Journal of Clinical Investigation 37~(4) (2007) 276--281.

\bibitem{wang2021learning}
S.~Wang, H.~Wang, P.~Perdikaris, Learning the solution operator of parametric
  partial differential equations with physics-informed deeponets, Science
  Advances 7~(40) (2021) eabi8605.

\bibitem{lu2021learning}
L.~Lu, P.~Jin, G.~Pang, Z.~Zhang, G.~E. Karniadakis, Learning nonlinear
  operators via deeponet based on the universal approximation theorem of
  operators, Nature Machine Intelligence 3~(3) (2021) 218--229.

\bibitem{raissi2019physics}
M.~Raissi, P.~Perdikaris, G.~E. Karniadakis, Physics-informed neural networks:
  A deep learning framework for solving forward and inverse problems involving
  nonlinear partial differential equations, Journal of Computational physics
  378 (2019) 686--707.

\bibitem{formaggia2003one}
L.~Formaggia, D.~Lamponi, A.~Quarteroni, One-dimensional models for blood flow
  in arteries, Journal of Engineering Mathematics 47 (2003) 251--276.

\bibitem{quarteroni2004mathematical}
A.~Quarteroni, L.~Formaggia, Mathematical modelling and numerical simulation of
  the cardiovascular system, Handbook of Numerical Analysis 12 (2004) 3--127.

\bibitem{xiao2014systematic}
N.~Xiao, J.~Alastruey, C.~Alberto~Figueroa, A systematic comparison between 1-d
  and 3-d hemodynamics in compliant arterial models, International Journal for
  Numerical Methods in Biomedical Engineering 30~(2) (2014) 204--231.

\bibitem{alastruey2011pulse}
J.~Alastruey, A.~W. Khir, K.~S. Matthys, P.~Segers, S.~J. Sherwin, P.~R.
  Verdonck, K.~H. Parker, J.~Peir{\'o}, Pulse wave propagation in a model human
  arterial network: assessment of 1-d visco-elastic simulations against in
  vitro measurements, Journal of Biomechanics 44~(12) (2011) 2250--2258.

\bibitem{saito2011one}
M.~Saito, Y.~Ikenaga, M.~Matsukawa, Y.~Watanabe, T.~Asada, P.-Y. Lagree,
  One-dimensional model for propagation of a pressure wave in a model of the
  human arterial network: Comparison of theoretical and experimental results,
  Journal of Biomechanical Engineering 133~(12) (2011).

\bibitem{olufsen2000numerical}
M.~S. Olufsen, C.~S. Peskin, W.~Y. Kim, E.~M. Pedersen, A.~Nadim, J.~Larsen,
  Numerical simulation and experimental validation of blood flow in arteries
  with structured-tree outflow conditions, Annals of Biomedical Engineering 28
  (2000) 1281--1299.

\bibitem{steele2003vivo}
B.~N. Steele, J.~Wan, J.~P. Ku, T.~J. Hughes, C.~A. Taylor, In vivo validation
  of a one-dimensional finite-element method for predicting blood flow in
  cardiovascular bypass grafts, IEEE Transactions on Biomedical Engineering
  50~(6) (2003) 649--656.

\bibitem{elad1991numerical}
D.~Elad, D.~Katz, E.~Kimmel, S.~Einav, Numerical schemes for unsteady fluid
  flow through collapsible tubes, Journal of Biomedical Engineering 13~(1)
  (1991) 10--18.

\bibitem{fullana2009branched}
J.-M. Fullana, S.~Zaleski, A branched one-dimensional model of vessel networks,
  Journal of Fluid Mechanics 621 (2009) 183--204.

\bibitem{wan2002one}
J.~Wan, B.~Steele, S.~A. Spicer, S.~Strohband, G.~R. Feijo{\'{}}~o, T.~J.
  Hughes, C.~A. Taylor, A one-dimensional finite element method for
  simulation-based medical planning for cardiovascular disease, Computer
  Methods in Biomechanics \& Biomedical Engineering 5~(3) (2002) 195--206.

\bibitem{sherwin2003computational}
S.~J. Sherwin, L.~Formaggia, J.~Peiro, V.~Franke, Computational modelling of 1d
  blood flow with variable mechanical properties and its application to the
  simulation of wave propagation in the human arterial system, International
  Journal for Numerical Methods in Fluids 43~(6-7) (2003) 673--700.

\bibitem{cavallini2008finite}
N.~Cavallini, V.~Caleffi, V.~Coscia, Finite volume and weno scheme in
  one-dimensional vascular system modelling, Computers \& Mathematics with
  Applications 56~(9) (2008) 2382--2397.

\bibitem{delestre2013well}
O.~Delestre, P.-Y. Lagr{\'e}e, A ‘well-balanced’finite volume scheme for
  blood flow simulation, International Journal for Numerical Methods in Fluids
  72~(2) (2013) 177--205.

\bibitem{matthys2007pulse}
K.~S. Matthys, J.~Alastruey, J.~Peir{\'o}, A.~W. Khir, P.~Segers, P.~R.
  Verdonck, K.~H. Parker, S.~J. Sherwin, Pulse wave propagation in a model
  human arterial network: assessment of 1-d numerical simulations against in
  vitro measurements, Journal of Biomechanics 40~(15) (2007) 3476--3486.

\bibitem{marchandise2009numerical}
E.~Marchandise, M.~Willemet, V.~Lacroix, A numerical hemodynamic tool for
  predictive vascular surgery, Medical Engineering \& Physics 31~(1) (2009)
  131--144.

\bibitem{mynard20081d}
J.~Mynard, P.~Nithiarasu, A 1d arterial blood flow model incorporating
  ventricular pressure, aortic valve and regional coronary flow using the
  locally conservative galerkin (lcg) method, Communications in Numerical
  Methods in Engineering 24~(5) (2008) 367--417.

\bibitem{wang20141d}
X.~Wang, 1d modeling of blood flow in networks: Numerical computing and
  applications, Ph.D. thesis, Paris 6 (2014).

\bibitem{li2020fourier}
Z.~Li, N.~Kovachki, K.~Azizzadenesheli, B.~Liu, K.~Bhattacharya, A.~Stuart,
  A.~Anandkumar, Fourier neural operator for parametric partial differential
  equations, arXiv preprint arXiv:2010.08895 (2020).

\bibitem{lu2021deepxde}
L.~Lu, X.~Meng, Z.~Mao, G.~E. Karniadakis, Deepxde: A deep learning library for
  solving differential equations, SIAM Review 63~(1) (2021) 208--228.

\bibitem{yu2018deep}
B.~Yu, et~al., The deep ritz method: a deep learning-based numerical algorithm
  for solving variational problems, Communications in Mathematics and
  Statistics 6~(1) (2018) 1--12.

\bibitem{lyu2022mim}
L.~Lyu, Z.~Zhang, M.~Chen, J.~Chen, {MIM}: A deep mixed residual method for
  solving high-order partial differential equations, Journal of Computational
  Physics 452 (2022) 110930.

\bibitem{yang2021local}
J.~Yang, Q.~Zhu, A local deep learning method for solving high order partial
  differential equations, Numerical Mathematics-Theory Methods and Applications
  (2021).

\bibitem{lu2021priori}
Y.~Lu, J.~Lu, M.~Wang, A priori generalization analysis of the deep ritz method
  for solving high dimensional elliptic partial differential equations, in:
  Conference on Learning Theory, PMLR, 2021, pp. 3196--3241.

\bibitem{muller2021error}
J.~M{\"u}ller, M.~Zeinhofer, Error estimates for the variational training of
  neural networks with boundary penalty, arXiv preprint arXiv:2103.01007
  (2021).

\bibitem{mishra2022estimates}
S.~Mishra, R.~Molinaro, Estimates on the generalization error of
  physics-informed neural networks for approximating a class of inverse
  problems for pdes, IMA Journal of Numerical Analysis 42~(2) (2022) 981--1022.

\bibitem{li2022generalization}
L.~Li, X.-C. Tai, J.~Yang, Generalization error analysis of neural networks
  with gradient based regularization, Communications in Computational Physics
  32~(4) (2022) 1007--1038.

\bibitem{li2022priori}
L.~Li, X.-c. Tai, J.~Yang, Q.~Zhu, Priori error estimate of deep mixed residual
  method for elliptic pdes, arXiv preprint arXiv:2206.07474 (2022).

\bibitem{he2016deep}
K.~He, X.~Zhang, S.~Ren, J.~Sun, Deep residual learning for image recognition,
  in: Proceedings of the IEEE Conference on Computer Vision and Pattern
  Recognition, 2016, pp. 770--778.

\bibitem{lu2022comprehensive}
L.~Lu, X.~Meng, S.~Cai, Z.~Mao, S.~Goswami, Z.~Zhang, G.~E. Karniadakis, A
  comprehensive and fair comparison of two neural operators (with practical
  extensions) based on fair data, Computer Methods in Applied Mechanics and
  Engineering 393 (2022) 114778.

\bibitem{li2021physics}
Z.~Li, H.~Zheng, N.~Kovachki, D.~Jin, H.~Chen, B.~Liu, K.~Azizzadenesheli,
  A.~Anandkumar, Physics-informed neural operator for learning partial
  differential equations, arXiv preprint arXiv:2111.03794 (2021).

\bibitem{goswami2022physics}
S.~Goswami, M.~Yin, Y.~Yu, G.~E. Karniadakis, A physics-informed variational
  deeponet for predicting crack path in quasi-brittle materials, Computer
  Methods in Applied Mechanics and Engineering 391 (2022) 114587.

\bibitem{westerhof2009arterial}
N.~Westerhof, J.-W. Lankhaar, B.~E. Westerhof, The arterial windkessel, Medical
  \& Biological Engineering \& Computing 47~(2) (2009) 131--141.

\bibitem{alastruey2012arterial}
J.~Alastruey, K.~H. Parker, S.~J. Sherwin, et~al., Arterial pulse wave
  haemodynamics, in: 11th International Conference on Pressure Surges, Vol.~30,
  Virtual PiE Led t/a BHR Group Lisbon, Portugal, 2012, pp. 401--443.

\bibitem{paviglianiti2022comparison}
A.~Paviglianiti, V.~Randazzo, S.~Villata, G.~Cirrincione, E.~Pasero, A
  comparison of deep learning techniques for arterial blood pressure
  prediction, Cognitive Computation 14~(5) (2022) 1689--1710.

\bibitem{lu2021physics}
L.~Lu, R.~Pestourie, W.~Yao, Z.~Wang, F.~Verdugo, S.~G. Johnson,
  Physics-informed neural networks with hard constraints for inverse design,
  SIAM Journal on Scientific Computing 43~(6) (2021) B1105--B1132.

\bibitem{avolio2009role}
A.~P. Avolio, L.~M. Van~Bortel, P.~Boutouyrie, J.~R. Cockcroft, C.~M. McEniery,
  A.~D. Protogerou, M.~J. Roman, M.~E. Safar, P.~Segers, H.~Smulyan, Role of
  pulse pressure amplification in arterial hypertension: experts’ opinion and
  review of the data, Hypertension 54~(2) (2009) 375--383.

\bibitem{gao2023failure}
Z.~Gao, L.~Yan, T.~Zhou, Failure-informed adaptive sampling for pinns, SIAM
  Journal on Scientific Computing 45~(4) (2023) A1971--A1994.

\bibitem{johnson2016mimic}
A.~E. Johnson, T.~J. Pollard, L.~Shen, L.-w.~H. Lehman, M.~Feng, M.~Ghassemi,
  B.~Moody, P.~Szolovits, L.~Anthony~Celi, R.~G. Mark, Mimic-iii, a freely
  accessible critical care database, Scientific Data 3~(1) (2016) 1--9.

\bibitem{misc_cuff-less_blood_pressure_estimation_340}
M.~Kachuee, M.~Kiani, M.~Hoda, M.~Shabany, {Cuff-Less Blood Pressure
  Estimation}, UCI Machine Learning Repository, {DOI}:
  https://doi.org/10.24432/C5B602 (2015).

\bibitem{hong2021deep}
J.~Hong, J.~Gao, Q.~Liu, Y.~Zhang, Y.~Zheng, Deep learning model with
  individualized fine-tuning for dynamic and beat-to-beat blood pressure
  estimation, in: 2021 IEEE 17th International Conference on Wearable and
  Implantable Body Sensor Networks (BSN), IEEE, 2021, pp. 1--4.

\bibitem{charlton2019modeling}
P.~H. Charlton, J.~Mariscal~Harana, S.~Vennin, Y.~Li, P.~Chowienczyk,
  J.~Alastruey, Modeling arterial pulse waves in healthy aging: a database for
  in silico evaluation of hemodynamics and pulse wave indexes, American Journal
  of Physiology-Heart and Circulatory Physiology 317~(5) (2019) H1062--H1085.

\bibitem{ieee2014ieee}
I.~S. Association, et~al., Ieee standard for wearable cuffless blood pressure
  measuring devices, IEEE Std (2014) 1708--2014.

\end{thebibliography}

\end{document}